\documentclass[table]{gtech}
\PassOptionsToPackage{table, usenames, dvipsnames}{xcolor}
\usepackage{xcolor}
\usepackage{amssymb}
\usepackage{multirow}
\usepackage{bigdelim}
\usepackage{longtable}
\usepackage{tabularray}
\usepackage{wrapfig}
\usepackage[most]{tcolorbox}
\usepackage{url}
\usepackage{float}
\usepackage{datatool}
\usepackage{enumitem}
\usepackage{subcaption} 
\usepackage[justification=centering]{caption}

\RequirePackage{tgpagella} 
\RequirePackage{mathpazo}  
\RequirePackage{inconsolata} 
\usepackage{makecell}

\usepackage{adjustbox}
\usepackage{tablefootnote}
\usepackage{threeparttable}

\usepackage{booktabs} 
\usepackage{array}    
\newcolumntype{C}[1]{>{\centering\arraybackslash}m{#1}}

\usepackage[utf8]{inputenc} 
\usepackage[T1]{fontenc}    
\usepackage{hyperref}       
\usepackage{url}            
\usepackage{booktabs}       
\usepackage{amsfonts}       
\usepackage{nicefrac}       
\usepackage{microtype}      
\usepackage{xcolor}         
\usepackage{xspace}
\usepackage{amsthm}   
\usepackage{amsmath}  
\usepackage{amssymb}  
\usepackage{bm}       

\theoremstyle{definition}

\renewcommand{\title}[1]{\newcommand{\titlelist}{{\huge\fontfamily{optimistic}\selectfont #1}}}

\newcommand{\ignore}[1]{}

\usepackage{arydshln}
\definecolor{CQColor}{rgb}{0.0,0.0,1.0} 

\usepackage{graphicx}
\usepackage{colortbl}
\usepackage{amssymb}
\usepackage{pifont}
\usepackage{booktabs,multirow}
\usepackage{makecell}
\usepackage{tabulary}
\usepackage{fontawesome5}
\usepackage{bbding}
\usepackage{multicol}

\newlength\savewidth

\title{
\huge{PowLU: An Activation Function for Stable Pre-Training of LLMs}
}

\author[]{Peijie Jiang}
\author[\dagger]{Yuqi Feng}
\author[]{Cunyin Peng}
\author[]{Qian Zhao}
\author[]{Jia Liu}
\author[]{KunLong Chen}
\author[*]{Zhiqiang Zhang}
\author[]{\\Jun Zhou}

\affiliation[]{Ling Team, Ant Group}
\contribution[*]{Corresponding author}
\contribution[\dagger]{Contribution during internship at Ant Group}

\abstract{
In contemporary large language models (LLMs), the swish-gated linear unit (SwiGLU) activation function is widely adopted to regulate the information flow and introduce non-linearity. 
For large positive inputs, SwiGLU approximates the quadratic function $x^2$, providing strong nonlinearity and expressive capacity. However, this property also causes numerical instability as the input or model scale increases, particularly in low-precision LLM training. The main reason is its approximate quadratic amplification, which enlarges the output range and exacerbates outliers. To address this issue, we propose a stable activation function, Power Linear Unit (PowLU), for large-scale LLM pre-training. Specifically, PowLU employs a rational power function to achieve adaptive nonlinearity, thereby improving representation ability and enabling stable training in spike regions. Moreover, we provide theoretical justification for several key properties of PowLU.
Scaling law experiments confirm that the performance is consistent across model sizes, and further experimental results with the Ling architecture (7.9B and 124B total parameters) demonstrate that PowLU achieves competitive results against SwiGLU and SwiGLU-Clip in large-scale training of LLMs. In addition, the experimental results also show that PowLU effectively improves the scalability of the large-scale training of LLMs.
}

\date{May 25, 2026\vspace{-1mm}}

\gtechdata[Correspondence]{\email{\{jiangpeijie.jpj,lingyao.zzq\}@antgroup.com}}

\begin{document}
\maketitle






\begin{figure*}[ht]
    \centering
    \subfloat[Activation Functions]{\includegraphics[width=0.45\linewidth]{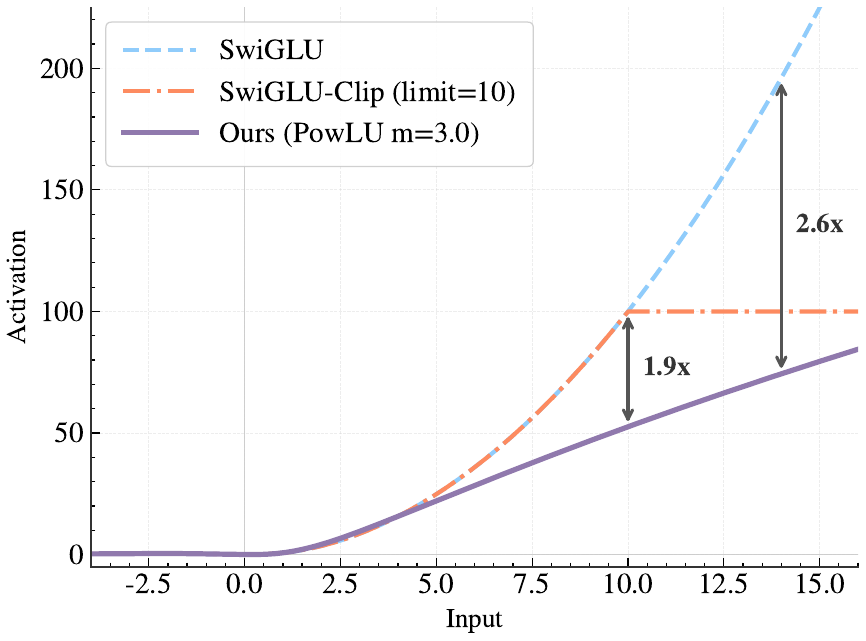}%
    \label{fig:plot_three_functions}}
    \hfil
    \subfloat[First-Order Derivatives of Functions]{\includegraphics[width=0.45\linewidth]{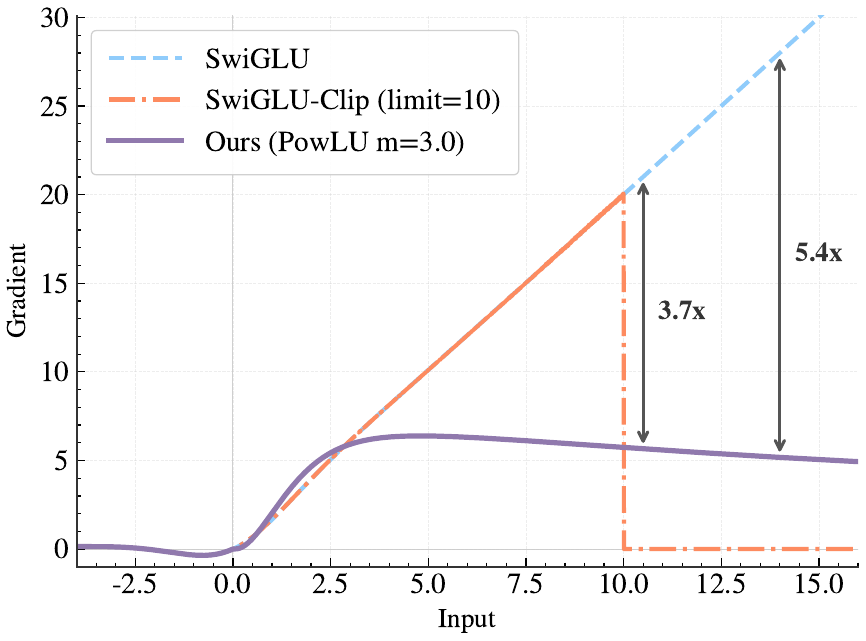}%
    \label{fig:plot_three_derivatives}}
    \caption{A visual comparison of SwiGLU, SwiGLU-Clip, and PowLU activation functions in terms of function curves and first-order derivatives. Both the activation output and the first derivative of SwiGLU are significantly larger than those of PowLU. 
    This gap continuously widens as the input increases, ultimately causing severe outliers.}
    \label{fig:plot_functions_derivatives}
\end{figure*}

\section{Introduction}

Large language models (LLMs) have emerged as a frontier in artificial intelligence research, demonstrating remarkable capabilities in natural language processing~\citep{team2023gemini,singh2025openai}. In the architecture of LLMs, the feed-forward network layers within Transformer blocks~\citep{vaswani2017attention} play a critical role in determining overall performance~\citep{mirzadeh2023relu,wei2024building}, primarily through the non-linearity enabled by the activation function between two linear transformation layers. Current studies and architectural designs indicate that the swish-gated linear unit (SwiGLU) activation function~\citep{shazeer2020glu} has become a prevalent choice in recent LLMs~\citep{grattafiori2024llama,yang2025qwen3}, as it can effectively enhance the model expressiveness and generalization.

\begin{figure*}
    \centering
    \subfloat[SwiGLU, Forward]{\includegraphics[width=0.24\linewidth]{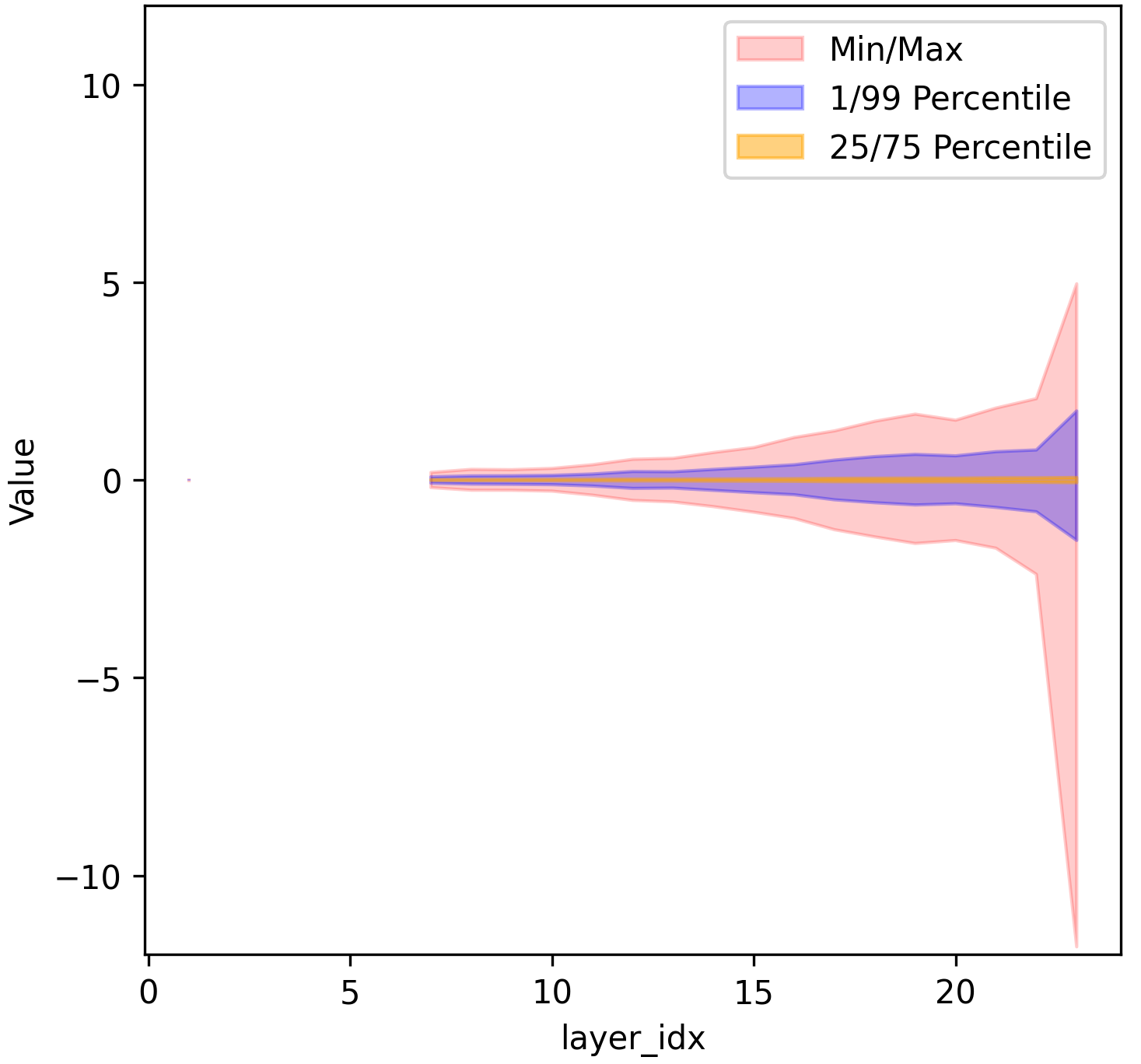}%
    \label{fig:baseline_linear_fc2_fwd_x_p99}}
    \hfil
    \subfloat[PowLU, Forward]{\includegraphics[width=0.24\linewidth]{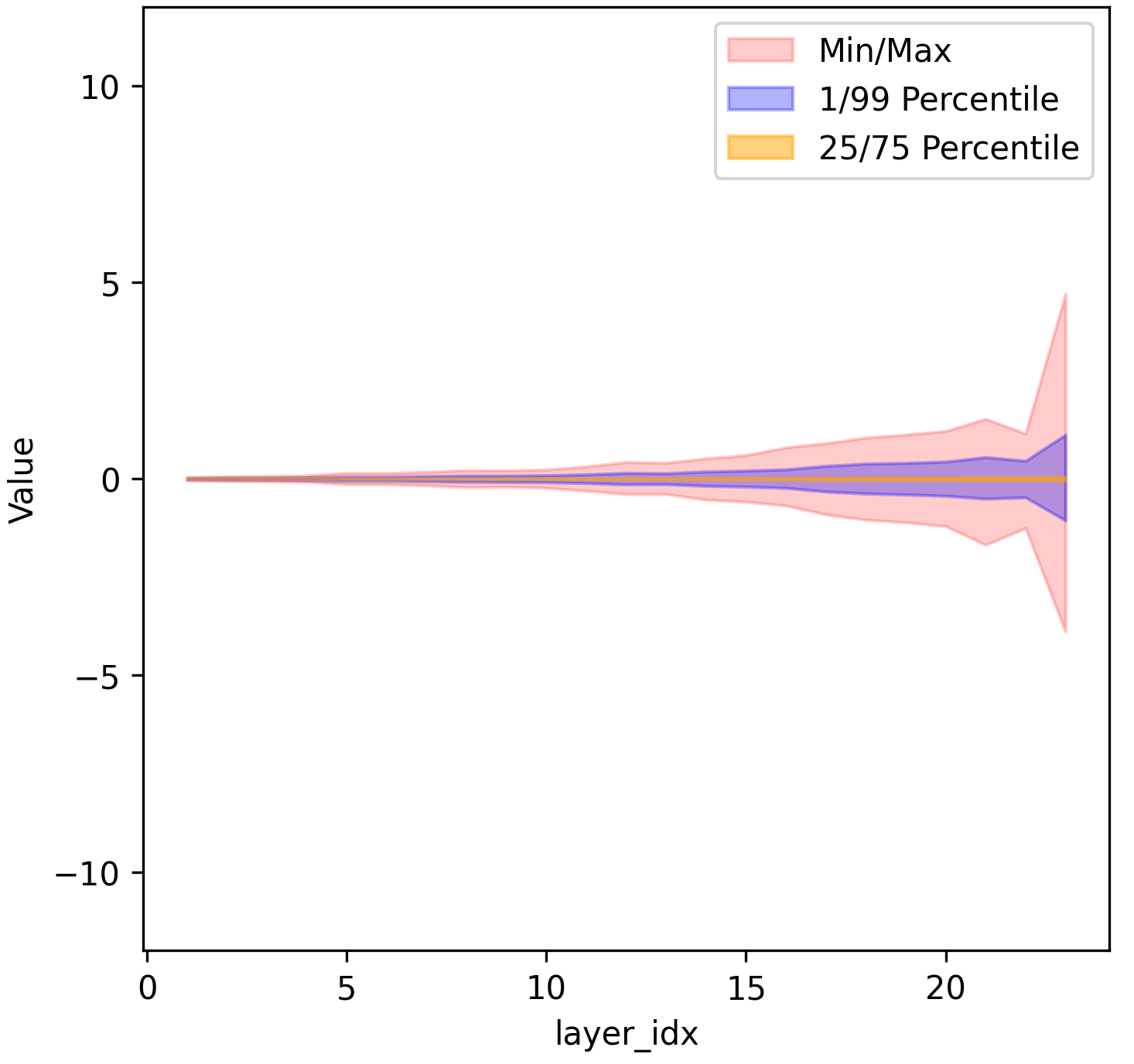}%
    \label{fig:mpau_linear_fc2_fwd_x_p99}}
    \hfil
    \subfloat[SwiGLU, Backward]{\includegraphics[width=0.24\linewidth]{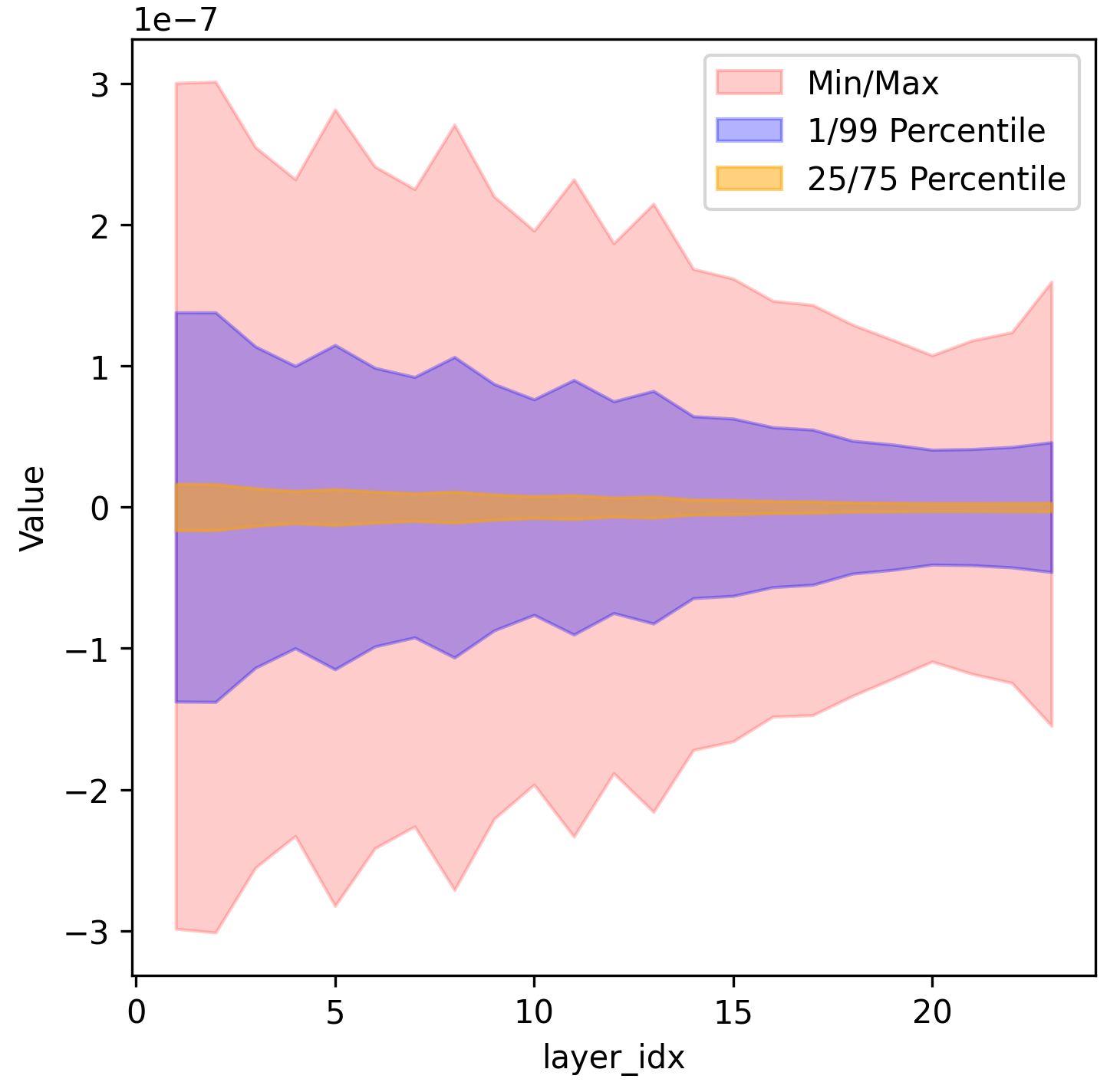}%
    \label{fig:baseline_linear_fc1_bwd_dy_p99}}
    \hfil
    \subfloat[PowLU, Backward]{\includegraphics[width=0.25\linewidth]{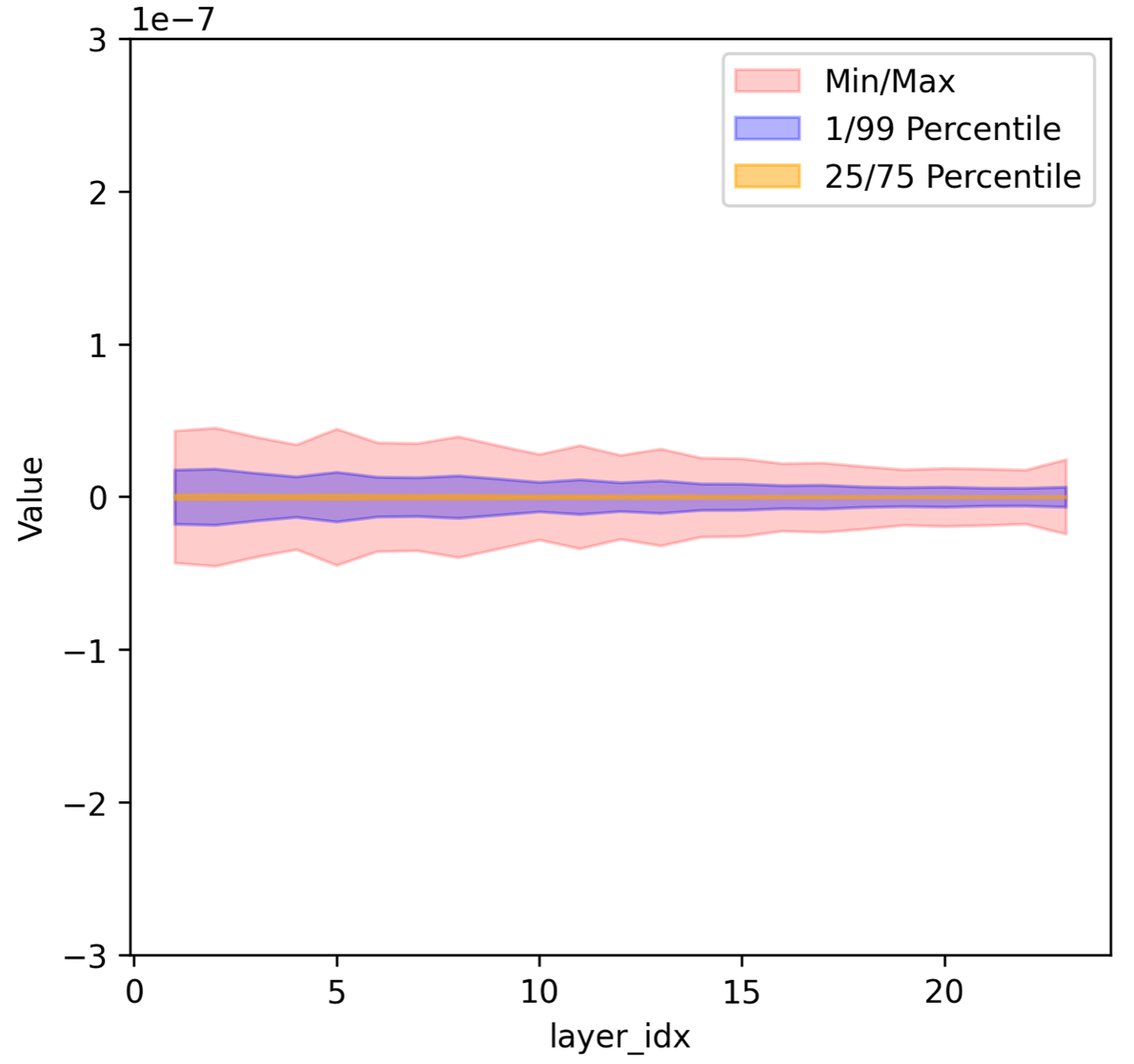}%
    \label{fig:mpau_linear_fc1_bwd_dy_p99}}
    \caption{Visualization of the P99 distribution in the pre-training of 7.9B MoE LLMs with SwiGLU and PowLU activation function. The results are recorded when 400B tokens are trained. The visualized layers are the linear layers in experts of the LLM. In the visualizations, the red band spans from the minimum to the maximum values to show the full range including extreme outliers. The purple band spans from P1 to P99 to capture 98\% of the distribution and indicate the effective dynamic range while excluding rare extremes. In addition, the orange band spans from P25 to P75 to represent the central 50\% of the distribution where most values concentrate.}
    \label{fig:vis_p99_experts}
\end{figure*}

Specifically, the SwiGLU activation function is formulated as the product of the input $x$ and $\text{SiLU}(x)$, where $\text{SiLU}(x)$ is formulated as $x \cdot \text{sigmoid}(x)$. According to the form of the SwiGLU activation function, the value of $\text{sigmoid}(x)$ approaches 1 as the input $x$ becomes large. As a result, the output of $\text{SwiGLU}(x)$ approaches $x^2$. The quadratic function $x^2$ has a strong amplifying effect on large inputs, resulting in a large number of outliers. Figure~\ref{fig:vis_p99_experts} shows the numerical distribution. As shown in Figure~\ref{fig:baseline_linear_fc2_fwd_x_p99}, the LLM with the SwiGLU activation function exhibits a wide red band which extends to large maximum values in the forward pass. Similarly, this phenomenon can also be observed in Figure~\ref{fig:baseline_linear_fc1_bwd_dy_p99}. These observations indicate that the SwiGLU activation function introduces outliers in the pre-training process. This problem will become much more severe due to the accumulation effect as the number of model layers increases, thus making the pre-training process unstable and prone to collapse. Furthermore, the stability of the pre-training can be further challenged if low-precision training techniques are adopted. This is because the low-precision training is more sensitive to the range of data distribution~\citep{chavan2024faster,lee2024fp8,hao2025low}, and the amplification effect of SwiGLU can cause some values to exceed the range of the corresponding precision, e.g., FP4 and FP8. As a result, the overall performance and stability of the training process will be limited.

To tackle this issue, we introduce a novel activation function called Power Linear Unit (PowLU) in this work. As shown in Figure~\ref{fig:plot_functions_derivatives}, PowLU can effectively limit the activation value when the input $x$ becomes large in the field of $x > 0$, while maintaining the non-linearity to ensure the representation ability of the whole LLM. Moreover, the form of PowLU is the same as that of SwiGLU for $x \le 0$. In PowLU, we introduce a rational power function of the square root of the input, to adaptively adjust the non-linearity level of the activation function. Furthermore, we also integrate the sigmoid function into PowLU to further enhance the degree of non-linearity. Owing to the above core designs, the rate of outliers is reduced compared with that of SwiGLU as shown in Figures~\ref{fig:mpau_linear_fc2_fwd_x_p99} and \ref{fig:mpau_linear_fc1_bwd_dy_p99}.

Our contributions can be summarized as follows: 
\begin{itemize}
    \item We introduce a stable activation function, i.e., PowLU, which can effectively limit the range of output data distribution when the positive input is large and ensure the non-linearity at the same time. As a result, the shortcoming of SwiGLU regarding the output range expanding can be effectively alleviated.
    \item We theoretically demonstrate the theoretical properties of the proposed PowLU activation function. Specifically, the continuity, differentiability, monotonicity, and bounded growth property are theoretically justified, thus supporting the rationale behind the design of the PowLU activation function.
    \item We have conducted scaling law experiments to demonstrate the consistent validity of PowLU. Furthermore, we also validate the effectiveness of PowLU in the large-scale pre-training for LLMs with 7.9 billion (B) and 124B parameters. The experimental results demonstrate that PowLU achieves competitive results compared with the widely used SwiGLU or SwiGLU-Clip activation function. Furthermore, the experimental results also show that PowLU can maintain the stability of the pre-training while reducing loss spikes.
\end{itemize}

\section{Related Works}

\subsection{Activation Functions}

The activation functions have played an essential role in ensuring the non-linearity of neural networks~\citep{kunc2024three}. Furthermore, they are also the key to ensuring the representation capability of Transformer-based models~\citep{minaee2024large}. Within this scope, the initial activation function in the transformer block is rectified linear unit (ReLU)~\citep{vaswani2017attention,nair2010rectified}. However, the ReLU activation function is not differentiable when the input is 0, leading to unstable model training. Focusing on this problem, the Gaussian error linear unit (GeLU)~\citep{hendrycks2016gaussian} is proposed and widely adopted in BERT~\citep{devlin-etal-2019-bert} and GPT~\citep{radford2018improving} models. Unfortunately, the activation functions discussed above all process each neuron independently without cross-feature mixing, thereby limiting expressive interactions between different hidden units~\citep{bai2022relu}. To achieve higher expressiveness of large-scale LLMs, researchers have introduced the gated linear unit (GLU) activation function with gating mechanism which adaptively controls the information flow~\citep{dauphin2017language,ramachandran2017searching}. Specifically, the gating mechanism is achieved by adopting a sigmoid gating function which selectively controls the information flow. Based on the GLU activation function, the SwiGLU activation function replaces the sigmoid gating function with the SiLU function~\citep{shazeer2020glu}, to boost the expressiveness and generalization ability of LLMs. However, the quadratic nature of SwiGLU can pose significant challenges for training stability of LLMs. In contrast, the PowLU activation function enhances the training stability by limiting the range of output data distribution while maintaining non-linearity.

\subsection{Stable Training of LLMs}

Training stability has become a critical aspect in the pre-training of LLMs, especially as model size continues to grow~\citep{takase2025spike,wortsman2024smallscale}. Training instability mainly manifests through loss spikes or gradient spikes, which can severely degrade model performance or even lead to training collapse~\citep{molybog2023theory,zhang2024when}. To address these stability challenges, researchers have proposed various techniques from different aspects, including optimization, architectural design, and activation function design. From the optimization perspective, gradient clipping has emerged as a fundamental technique for preventing gradient explosions and improving training stability~\citep{pascanu2013difficulty}. For example, the introduction of spike-aware Adam optimizer with momentum reset represents a significant advancement, employing momentum reset and spike-aware gradient clipping to counteract gradient spikes~\citep{huang2025spam}. Furthermore, the stable variant of the above method extends the method with adaptive spike recognition thresholds and dynamic scaling strategies, enabling more stable low-precision training process~\citep{huang2025stablespam}. From the architectural design perspective, residual connections and layer normalization have become standard components in modern LLMs, mitigating the vanishing and exploding gradient problems~\citep{vaswani2017attention}. Recent innovations such as attention residuals replace the fixed accumulation of layer outputs with softmax attention over preceding layers, yielding more uniform output magnitudes and gradient distribution across depth~\citep{team2026attention}. Beyond the two directions discussed above, activation function design offers another promising way to improve training stability. For example, SwiGLU-Clip~\citep{agarwal2025gpt} improves stability by clamping the linear component and capping the gate component, which effectively suppresses activation outliers. However, such hard truncation may discard useful information once activations exceed the predefined thresholds. To address this issue, we introduce the PowLU activation function, which suppresses activation growth more smoothly while better preserving representational capacity.

\section{PowLU Activation Function}

\subsection{Formulation}

\textbf{Function Expression.} The formulation of the proposed PowLU activation function is presented as Eq.~(\ref{eq:mpau_formulation}):
\begin{equation}\label{eq:mpau_formulation}
\text{PowLU}(x) =
\begin{cases}
    x \cdot x^{\frac{m}{\sqrt{x} + 1}} \cdot \text{sigmoid}(x) \quad &x > 0 \\
    x^2 \cdot \text{sigmoid}(x) \quad &x \leq 0
\end{cases}
\end{equation}
where $x$ denotes the input, $m$ denotes the hyperparameter in the range of $0 < m < 10$, and $\text{sigmoid}(x) = 1 / (1 + e^{-x})$ denotes the sigmoid function which ensures the non-linearity of the PowLU activation function. Moreover, the format of $\text{PowLU}(x)$ is the same as that of SwiGLU activation function when $x \leq 0$. 

In our training process, the PowLU activation function is implemented as $\text{PowLU}(x_1, x_2) = x_1 \cdot f(x_2)$, where $x_1$ and $x_2$ are two linear projections of the same input $x$. When $x_2 > 0$, the function $f(x_2)$ is implemented as $f(x_2) = x_2^{m / (\sqrt{x_2} + 1)} \cdot \text{sigmoid}(x_2)$. Furthermore, when $x_2 \leq 0$, the implementation of $f(x_2)$ follows that of the SiLU function.

\textbf{Motivations.} In the case of $x > 0$, $\sqrt{x}$ is included to slow down the rate at which the activation function degenerates into a linear form as $x$ increases. To be more specific, when $x$ is increasing, $m / (\sqrt{x} + 1)$ approaches 0 and $\text{sigmoid}(x)$ approaches 1, and the PowLU activation function is approximately equal to $x$. In this case, $\sqrt{x}$ can decrease the speed at which $m / (\sqrt{x} + 1)$ approaches 0, thus reducing the speed at which the activation function degenerates to $x$. Meanwhile, $\sqrt{x}$ can further enhance the non-linearity of the PowLU activation function compared with $x$. 

In addition, term $\sqrt{x} + 1$ is designed to ensure the feasibility of differentiation when $x$ approaches $0^{+}$. Specifically, term $m / (\sqrt{x} + 1)$ approaches $m$ as $x$ approaches $0^{+}$, and the differentiation can be guaranteed. In another case where term $m / (\sqrt{x} + 1)$ is changed to $m / \sqrt{x}$, its value approaches $+\infty$ as $x$ approaches $0^{+}$, thus the differentiability cannot be guaranteed. Therefore, we add a constant 1 to the denominator to avoid this problem.

\subsection{Theoretical Analysis}
From the visualizations shown in Figure~\ref{fig:plot_functions_derivatives}, we can observe that PowLU has the monotonically increasing property, continuity, differentiability, and bounded growth property when the hyperparameter $m$ is set to 3. In this subsection, we theoretically analyze the above properties of PowLU. The detailed analysis is presented as follows.

\textbf{Continuity.} The PowLU activation function is composed of elementary functions when $x \neq 0$. Therefore, it is continuous in this case. When $x = 0$, the function value $\text{PowLU}(0) = 0$. Meanwhile, the left-hand limit $\lim_{x \rightarrow 0^{-}} \text{PowLU}(x) = 0^2 \cdot 1 / (1 + e^0) = 0$. Moreover, the right-hand limit can be calculated as $\lim_{x \rightarrow 0^{+}} 0^{1+ m} \cdot 1 / 2$. Because the hyperparameter $m$ satisfies $m > 0$, the right-hand limit is 0. Since the left-hand limit, the function value, and the right-hand limit are all equal to 0, the PowLU activation function is continuous at $x = 0$. 

\textbf{Differentiability.} Because the PowLU activation function is composed of differentiable elementary functions when $x \neq 0$, it is also differentiable in these cases. When $x = 0$, we analyze values of left-hand and right-hand derivatives to show the differentiability. Specifically, the left-hand derivative can be written as $\text{PowLU}'_{-}(0) = \lim_{h \rightarrow 0^{-}} (\text{PowLU}(h) - \text{PowLU}(0)) / h = \lim_{h \rightarrow 0^{-}} h \cdot \text{sigmoid}(h) = 0$. Furthermore, the right-hand derivative can be written as $\text{PowLU}'_{+}(0) = \lim_{h \rightarrow 0^{+}} (\text{PowLU}(h) - \text{PowLU}(0)) / h = \lim_{h \rightarrow 0^{+}} h^{m / (\sqrt{h} + 1)} \cdot \text{sigmoid}(h)$. Because $m > 0$, $\text{PowLU}'_{+}(0)$ is equal to 0. Since the left-hand derivative is equal to the right-hand derivative, the PowLU activation function is differentiable at $x = 0$. 

\textbf{Monotonicity.} In the case of $x > 0$, PowLU is monotonically increasing when the hyperparameter $m$ satisfies $0 < m < 10$. The detailed analysis can be found in Appendix~\ref{subsec:monotonicity}.

\textbf{Bounded Growth Property.} The bounded growth property is analyzed based on the behavior of the PowLU activation function as $x \rightarrow -\infty$ and $x \rightarrow +\infty$. In the case of $x \rightarrow -\infty$, PowLU has properties consistent with SwiGLU~\citep{shazeer2020glu} and converges to 0 in this direction. In the case of $x \rightarrow +\infty$, the limit of the PowLU activation function can be written as $\lim_{x \rightarrow +\infty} \text{PowLU}(x) = \lim_{x \rightarrow +\infty} x^{1 + m / (\sqrt{x} + 1)} \cdot \text{sigmoid}(x)$. As $x \rightarrow +\infty$, term $\text{sigmoid}(x) \to 1$, and the exponent part $1 + m / (\sqrt{x} + 1) \rightarrow 1$. As a result, the PowLU activation function's behavior approximates $\lim_{x \to +\infty} x^1 \cdot 1 = +\infty$. Therefore, the PowLU activation function is unbounded in the positive direction and exhibits approximately linear growth, but the growth rate is still lower than that of the initial SwiGLU activation function, i.e., the square of $x$.

\section{Experiments}

\subsection{Experimental Setup}

In this section, the experimental setup is introduced from two aspects, i.e., training settings and evaluation benchmarks.

\subsubsection{Training Settings}

To evaluate the performance of different activation functions (i.e., SwiGLU~\citep{shazeer2020glu}, SwiGLU-Clip~\citep{agarwal2025gpt}, and PowLU), we place them between two linear layers in experts and shared experts in Mixture-of-Experts (MoE) models following the Ling architecture~\citep{team2025every}. The LLMs are of different model sizes. In particular, we have adopted a series of LLMs with a small number of parameters for scaling law experiments. Meanwhile, we have also trained larger models with 7.9B and 124B parameters for a large number of GPU hours, to demonstrate the effectiveness of PowLU on large-scale LLMs. Furthermore, we have also performed ablation studies with the LLMs adopted for the scaling law experiments. As for the hyperparameter $m$, we choose the value 3.0 in our experiments, and the parameter studies of $m$ can be found in Section~\ref{subsubsec:parameter_study}.

\subsubsection{Evaluation Benchmarks}

Following the conventions~\citep{team2025every}, we choose evaluation benchmarks from three categories, i.e., World Knowledge, Language \& Reasoning, and Math \& Code to evaluate LLMs with different activation functions. The benchmarks adopted are listed as follows.
\begin{itemize}
    \item \textbf{World Knowledge.} The evaluation benchmarks for experiments contain AGIEval~\citep{zhong2024agieval}, MMLU~\citep{hendrycks2021measuring}, MMLU-Pro~\citep{wang2024mmlu}, MMMLU~\citep{wang2024mmmlu}, C-Eval~\citep{huang2023c}, CMMLU~\citep{li2024cmmlu}, SuperGPQA~\citep{du2025supergpqa}, TriviaQA~\citep{joshi2017triviaqa}, and ARC-challenge~\citep{clark2018think}.
    \item \textbf{Language \& Reasoning.} The evaluation benchmarks are BBH~\citep{suzgun2023challenging}, HellaSwag~\citep{zellers2019hellaswag}, and WinoGrande~\citep{sakaguchi2021winogrande}.
    \item \textbf{Math \& Code.} The evaluation benchmarks are HumanEval~\citep{chen2021evaluating}, GSM8K~\citep{cobbe2021training}, MATH~\citep{hendrycks2021math}, MGSM~\citep{shi2023language}, and CMATH~\citep{wei2023cmath}.
\end{itemize}

\subsection{Main Results}

The main results come from three aspects, i.e., the scaling law experiments and the pre-train results of LLMs with 7.9B and 124B parameters. The experimental results and corresponding analysis are detailed as follows.

\subsubsection{Scaling Law Experiments}

\begin{figure*}[ht]
    \centering
    \includegraphics[width=0.65\linewidth]{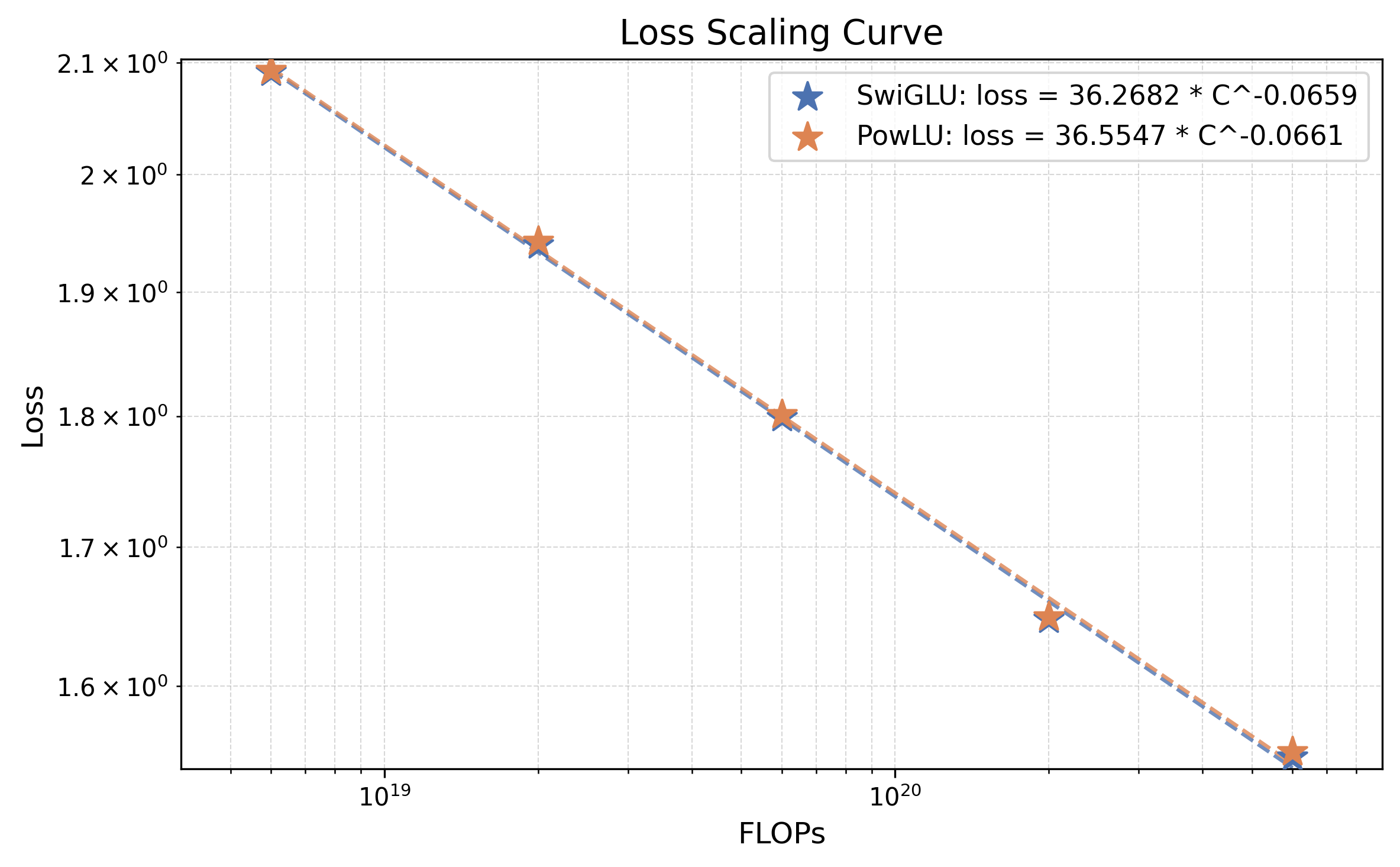}
    \caption{The fitted scaling law curves for SwiGLU and PowLU.}
    \label{fig:scaling_law}
\end{figure*}

We have conducted scaling law experiments on a series of MoE LLMs. The details of LLMs and training settings are presented in Table~\ref{tab:scaling_law}. Please note that the corresponding settings for LLMs with SwiGLU and PowLU activation functions remain consistent. As shown in Figure~\ref{fig:scaling_law}, the loss scaling curves of SwiGLU and PowLU almost overlap, indicating that SwiGLU and PowLU have roughly the same performance on MoE LLMs with 26 million (M) to 368M activated parameters.

\begin{table*}[ht]
    \caption{Model configurations and hyperparameters for scaling law experiments. ``Act. Params.'' refers to the number of activated parameters in MoE models. All models are trained with the sequence length of 4,096.}
    \label{tab:scaling_law}
    \centering
    \begin{tabular}{ccccc}
    \toprule
    \textbf{Act. Params.} & \textbf{Layer} & \textbf{Hidden} & \textbf{lr} & \textbf{batch size} \\
    \midrule
    26M & 10 & 512 & 0.00156 & 128 \\
    47M & 12 & 640 & 0.0013 & 192 \\
    92M & 16 & 768 & 0.0011 & 256 \\
    199M & 20 & 1024 & 0.00091 & 448 \\
    368M & 24 & 1280 & 0.00077 & 640 \\
    \bottomrule
    \end{tabular}
\end{table*}

\subsubsection{Results on 7.9B Model}

In this set of experiments, the LLMs with 7.9B parameters in total are pre-trained with 600B tokens. After that, the LLM with the proposed PowLU activation function is compared with other two baselines, i.e., LLMs with SwiGLU and SwiGLU-Clip activation functions. The experimental results are presented in Table~\ref{tab:main_tiny}. These results demonstrate that PowLU achieves competitive performance on the 17 evaluation benchmarks selected. To be more specific, PowLU enhances training stability while retaining desirable non-linear transformation properties, thus effectively preserving the model's expressive capacity. This set of results further confirms the effectiveness of PowLU.

\begin{table*}[ht]
    \caption{Comparisons among SwiGLU, SwiGLU-Clip, and PowLU activation functions. The LLM adopted for pre-training is an MoE model with 7.9B total parameters. The highest results are in bold font and the second highest results are underlined.}
    \label{tab:main_tiny}
    \centering
    \begin{tabular}{clccc}
    \toprule
    \textbf{Types} & \textbf{Benchmarks} & \textbf{SwiGLU} & \textbf{SwiGLU-Clip} & \textbf{PowLU} \\
    \midrule
    & Trained Tokens & 600B & 600B & 600B \\
    \midrule
    \multirow{9}{*}{World Knowledge} & AGIEval & \textbf{31.75} & 30.23 & \underline{31.13} \\
    & MMLU & 53.95 & \underline{54.12} & \textbf{54.92} \\
    & MMLU-Pro & 21.56 & \underline{21.79} & \textbf{24.00} \\
    & MMMLU & 31.77 & \underline{32.40} & \textbf{32.61} \\
    & C-Eval & 51.99 & \underline{52.62} & \textbf{52.96} \\
    & CMMLU & \underline{52.69} & 50.24 & \textbf{52.82} \\
    & SuperGPQA & \textbf{17.67} & \underline{17.14} & 17.02 \\
    & TriviaQA & 47.86 & \textbf{48.87} & \underline{48.18} \\
    & ARC-challenge & 51.53 & \underline{51.86} & \textbf{55.93} \\
    \midrule
    \multirow{3}{*}{Language \& Reasoning} & BBH & \underline{38.82} & 38.22 & \textbf{38.96} \\
    & HellaSwag & 66.24 & \underline{66.31} & \textbf{66.46} \\
    & WinoGrande & 63.14 & \underline{64.56} & \textbf{65.90} \\
    \midrule
    \multirow{5}{*}{Math \& Code} & HumanEval & \underline{25.61} & 23.17 & \textbf{26.83} \\
    & GSM8K & 30.40 & \underline{32.30} & \textbf{33.74} \\
    & MATH & \underline{22.98} & 22.64 & \textbf{23.98} \\
    & MGSM & 15.93 & \underline{17.40} & \textbf{18.40} \\
    & CMATH & \underline{63.21} & \textbf{63.39} & 63.11 \\
    \bottomrule
    \end{tabular}
\end{table*}

\subsubsection{Results on 124B Model}

To validate the effectiveness of PowLU on larger-scale LLMs, we train LLMs with 124B parameters in total with 800B tokens. The LLMs are integrated with SwiGLU or PowLU activation functions for comparison. The experimental results are presented in Table~\ref{tab:main_flash}. These results demonstrate that the PowLU-based LLM delivers competitive overall performance. The above observation further confirms that integrating the PowLU activation function can effectively maintain the performance of the larger-scale LLM while enhancing the training stability.

\begin{table*}[ht]
    \caption{Comparisons among SwiGLU and PowLU activation functions. The LLM adopted for pre-training is an MoE model with 124B total parameters. The highest results are in bold font.}
    \label{tab:main_flash}
    \centering
    \begin{tabular}{clcc}
    \toprule
    \textbf{Types} & \textbf{Benchmarks} & \textbf{SwiGLU} & \textbf{PowLU} \\
    \midrule
    & Trained Tokens & 800B & 800B \\
    \midrule
    \multirow{9}{*}{World Knowledge} & AGIEval & 53.03 & \textbf{53.75} \\
    & MMLU & 69.10 & \textbf{69.14} \\
    & MMLU-Pro & \textbf{40.75} & 40.12 \\
    & MMMLU & 46.27 & \textbf{48.10} \\
    & C-Eval & \textbf{71.74} & 71.56 \\
    & CMMLU & \textbf{71.58} & 71.41 \\
    & SuperGPQA & 25.93 & \textbf{26.16} \\
    & TriviaQA & 66.31 & \textbf{66.91} \\
    & ARC-challenge & 77.29 & \textbf{83.05} \\
    \midrule
    \multirow{3}{*}{Language \& Reasoning} & BBH & 62.07 & \textbf{63.36} \\
    & HellaSwag & 76.23 & \textbf{76.24} \\
    & WinoGrande & \textbf{75.45} & 73.72 \\
    \midrule
    \multirow{5}{*}{Math \& Code} & HumanEval & 54.27 & \textbf{55.49} \\
    & GSM8K & \textbf{70.81} & 69.90 \\
    & MATH & 42.22 & \textbf{44.98} \\
    & MGSM & 54.00 & \textbf{54.80} \\
    & CMATH & 80.69 & \textbf{83.33} \\
    \bottomrule
    \end{tabular}
\end{table*}

\subsection{Stability Evaluation}

In this subsection, we evaluate the stability of PowLU in terms of the loss spike, the numerical distribution of tensors, and the outlier channels.

\subsubsection{Loss Spike Analysis}

\begin{figure*}[ht]
    \centering
    \includegraphics[width=\linewidth]{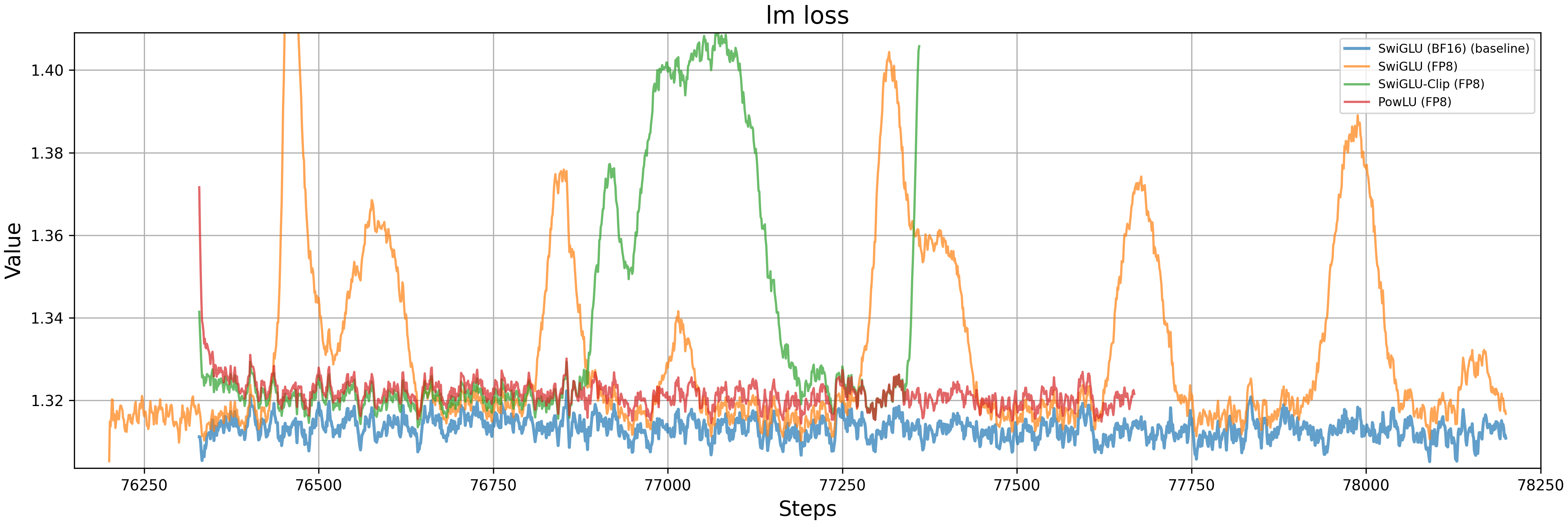}
    \caption{Visualization of the training loss curves of large-scale LLMs with SwiGLU, SwiGLU-Clip, or PowLU activation functions. \textbf{Blue Line:} SwiGLU-based LLM trained with BF16 precision. \textbf{Orange Line:} SwiGLU-based LLM trained with FP8 precision. \textbf{Green Line:} SwiGLU-Clip-based LLM trained with FP8 precision. \textbf{Red Line:} PowLU-based LLM trained with FP8 precision.}
    \label{fig:loss_spike}
\end{figure*}

In the training processes of larger-scale LLMs with SwiGLU or SwiGLU-Clip activation functions, we have faced the loss spike problem which makes the training process unstable. In contrast, the proposed PowLU activation function shows improved stability in this case. In this part, we have visualized the loss curves of the above training processes in Figure~\ref{fig:loss_spike} for further analysis.

In our experiments, loss spikes mainly appear after 76,200 training steps. As shown in Figure~\ref{fig:loss_spike}, the visualized loss curves demonstrate a divergence in stability. Specifically, the SwiGLU-based LLM trained with BF16 precision (blue line) achieves the relatively low loss values and stable training process. This phenomenon mainly comes from two aspects. First, the training precision of this experiment (BF16) is higher than that of other experiments (FP8), thus the loss values of this experiment are lower than those of others. Second, the experiments for SwiGLU-Clip and PowLU are conducted via replacing the SwiGLU activation function after training certain steps. As a result, there is a loss recovery process and the loss values are higher. Moreover, the SwiGLU-Clip activation function (green line) delayed the timing of loss spikes compared to the SwiGLU baseline with FP8 training (orange line), but it also suffers from the loss spike around training step 77,000. In addition, the proposed PowLU activation function (red line) maintains a consistently smooth and low loss trajectory, hovering around 1.32 without significant deviations. The above empirical observations indicate that the proposed PowLU activation function effectively mitigates the occurrence of loss spikes, enhancing the training stability of large-scale LLMs compared to commonly used SwiGLU and SwiGLU-Clip activation functions.

\subsubsection{Numerical Distribution of Tensors}

\begin{figure*}[ht]
    \centering
    \subfloat[SwiGLU, Forward]{\includegraphics[width=0.24\linewidth]{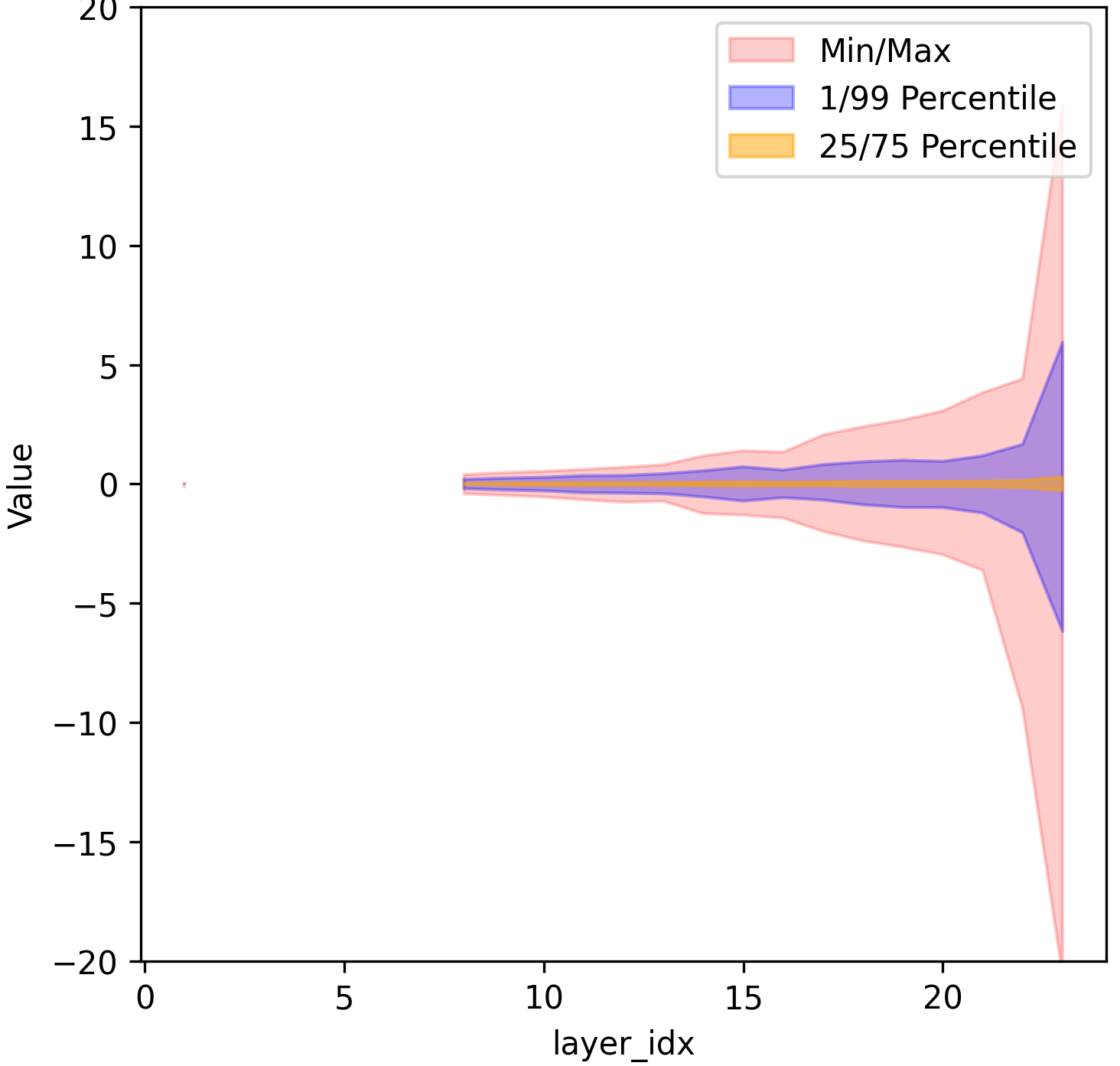}%
    \label{fig:baseline_shared_experts_forward}}
    \hfil
    \subfloat[PowLU, Forward]{\includegraphics[width=0.24\linewidth]{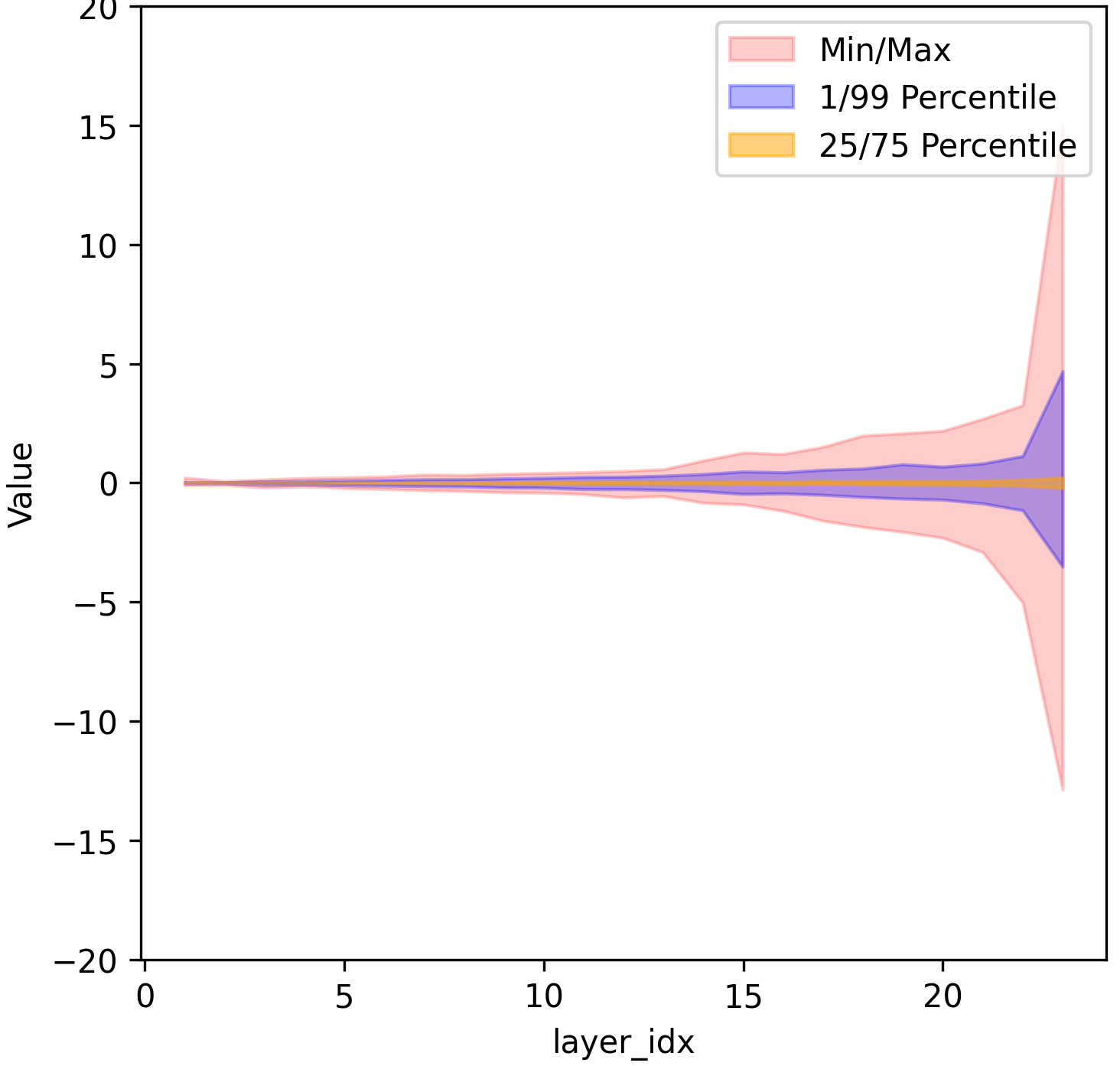}%
    \label{fig:new_act_shared_experts_forward}}
    \hfil
    \subfloat[SwiGLU, Backward]{\includegraphics[width=0.24\linewidth]{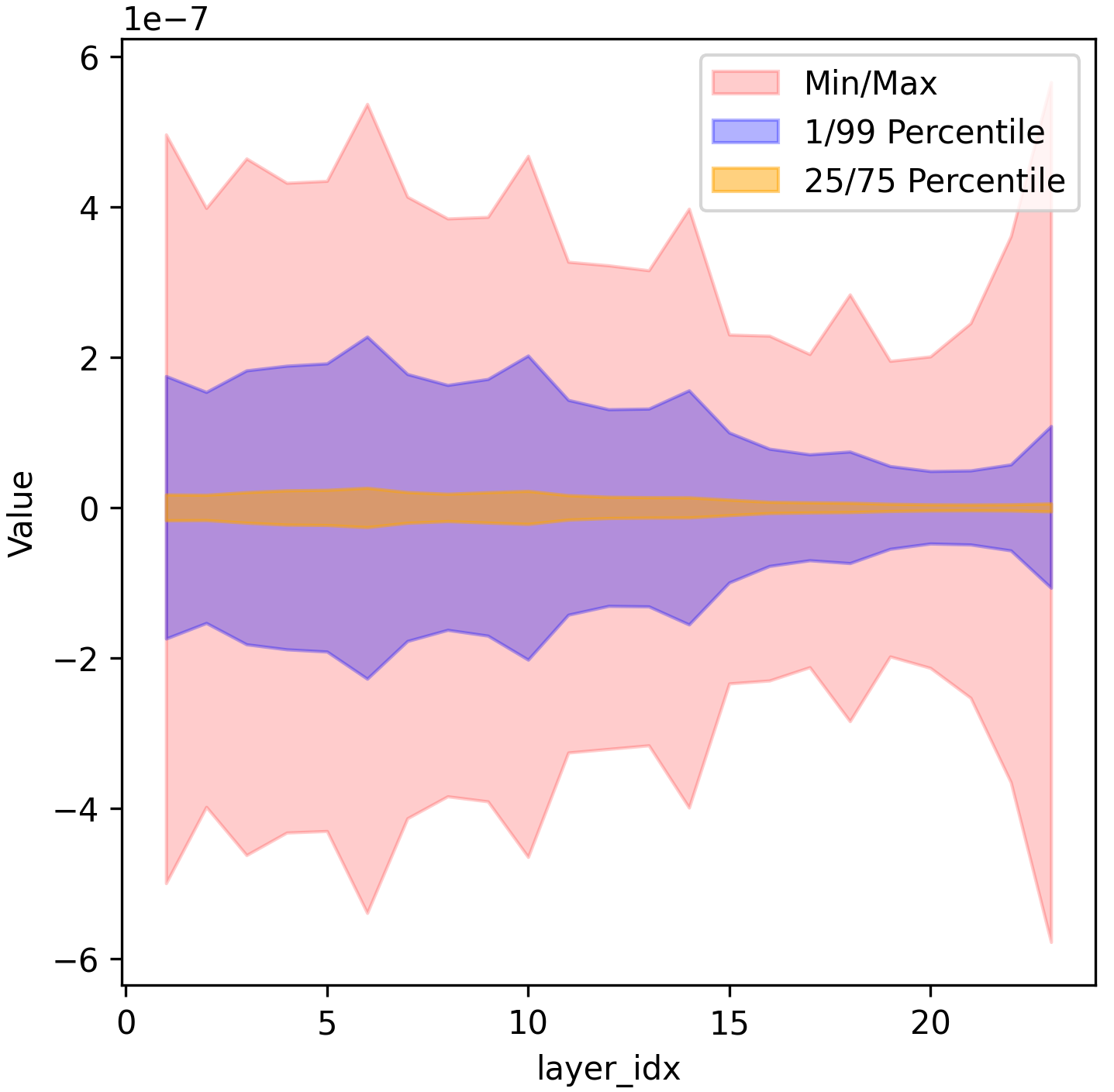}%
    \label{fig:baseline_shared_experts_backward}}
    \hfil
    \subfloat[PowLU, Backward]{\includegraphics[width=0.25\linewidth]{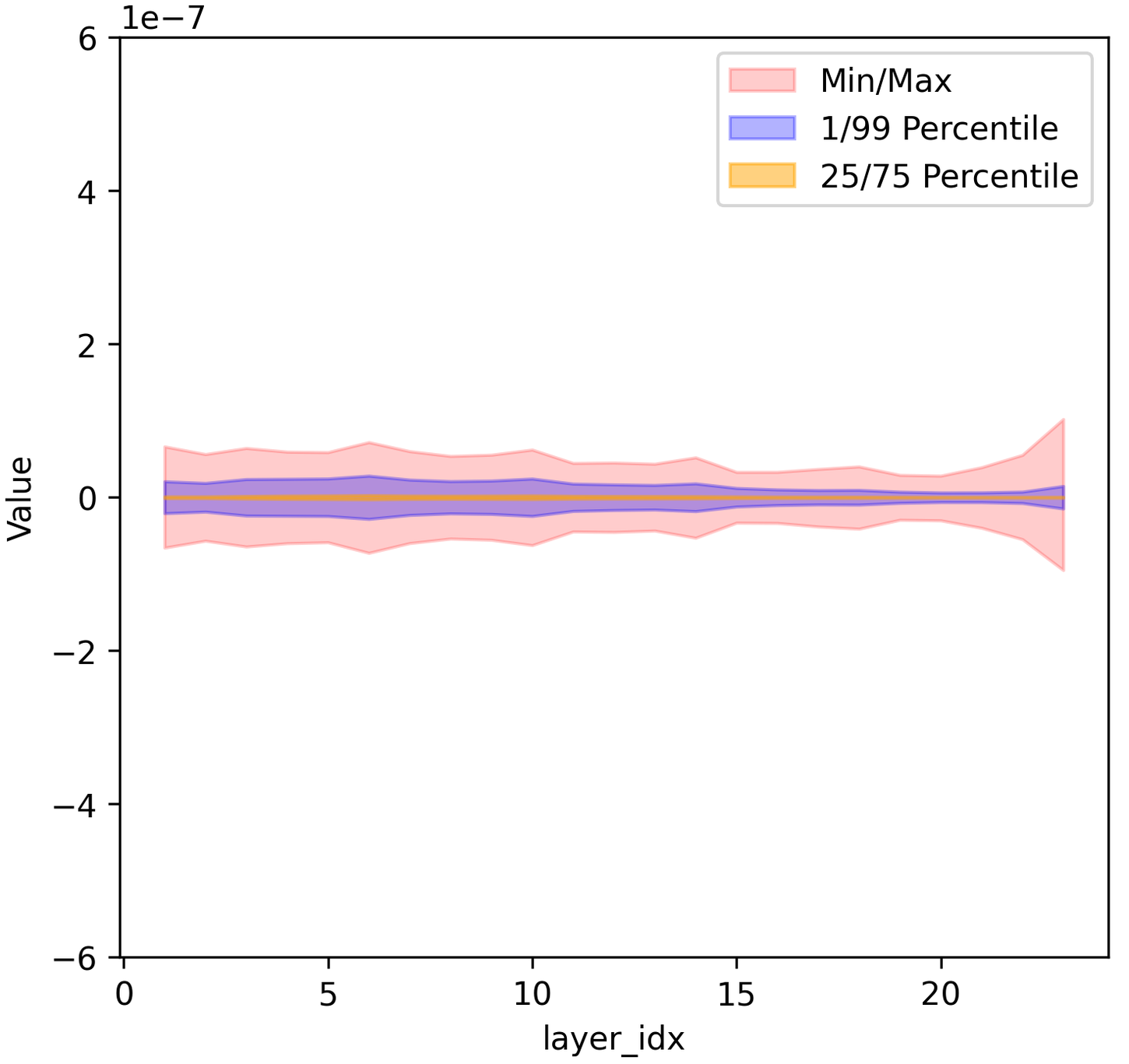}%
    \label{fig:new_act_shared_experts_backward}}
    \caption{Visualization of the P99 distribution in the low-precision training of 7.9B MoE LLMs with SwiGLU activation function or PowLU activation function. The results are recorded when 400B tokens are trained. The visualized layers are the linear layers in the shared experts of the LLM.}
    \label{fig:vis_p99_shared_experts}
\end{figure*}

We introduce the distribution plots to monitor the numerical distribution of tensors across layers, thus diagnosing training instabilities. Similar to Figure~\ref{fig:vis_p99_experts}, we collect six percentile-based statistics of two tensor types for linear layers in shared experts. Specifically, we record the input activation tensor in the forward pass of the second linear layer after the activation function, and the gradient tensor in the backward pass of the first linear layer before the activation function. The visualizations are presented in Figure~\ref{fig:vis_p99_shared_experts}.

As shown in Figure~\ref{fig:vis_p99_shared_experts}, SwiGLU (Figure~\ref{fig:baseline_shared_experts_forward}) exhibits a wider red band which extends to much higher maximum values compared to PowLU (Figure~\ref{fig:new_act_shared_experts_forward}) in the forward pass. This trend becomes more obvious in the backward pass as shown in Figures~\ref{fig:baseline_shared_experts_backward} and \ref{fig:new_act_shared_experts_backward}. The gradients of the LLM with SwiGLU display a highly volatile and expansive red band, demonstrating the presence of severe gradient outliers which can destabilize the training process. In contrast, the backward pass of the LLM with PowLU demonstrates a more constrained red band with lower maximum values, suggesting that the proposed PowLU activation function effectively suppresses extreme values. In summary, the PowLU activation function can effectively maintain a tighter range between the minimum and maximum values across both forward and backward propagation, thus mitigating the occurrence of outliers and ensuring a more stable numerical distribution for large-scale pre-training.

\subsubsection{Outlier Channels Analysis}

\begin{figure*}[ht]
    \centering
    \subfloat[SwiGLU, Forward]{\includegraphics[width=0.24\linewidth]{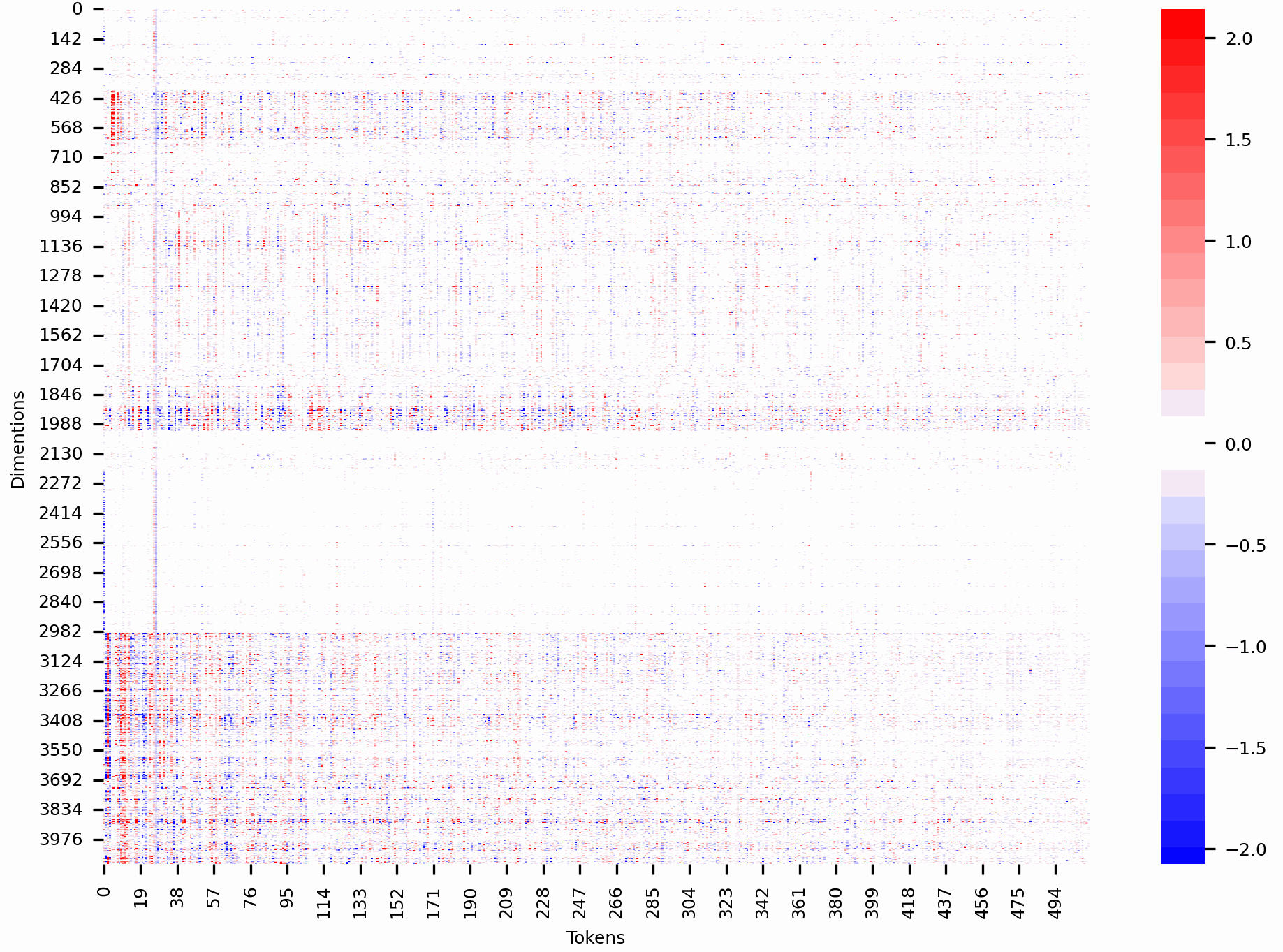}%
    \label{fig:channels_baseline_experts_forward}}
    \hfil
    \subfloat[PowLU, Forward]{\includegraphics[width=0.24\linewidth]{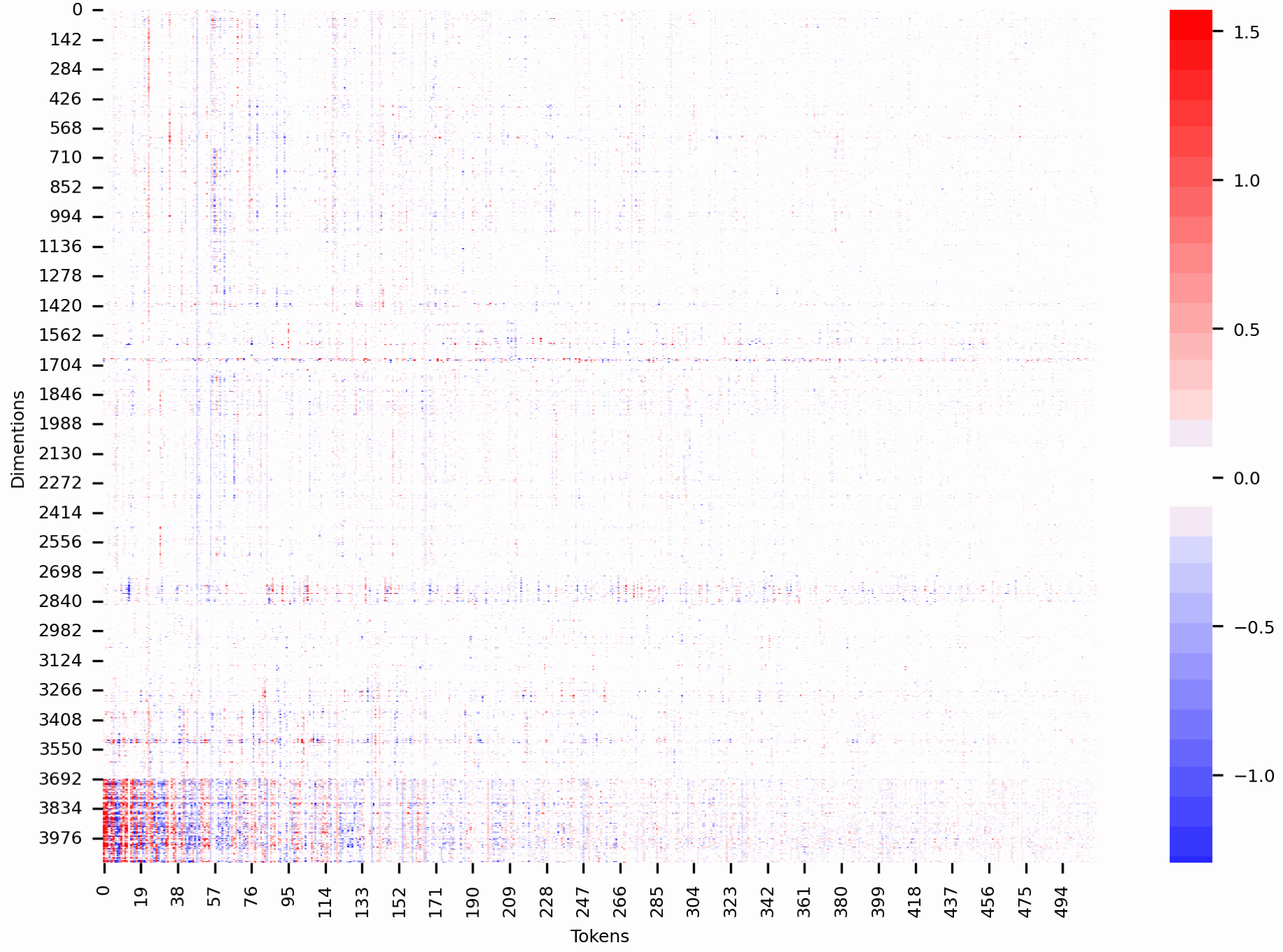}%
    \label{fig:channels_new_act_experts_forward}}
    \hfil
    \subfloat[SwiGLU, Backward]{\includegraphics[width=0.24\linewidth]{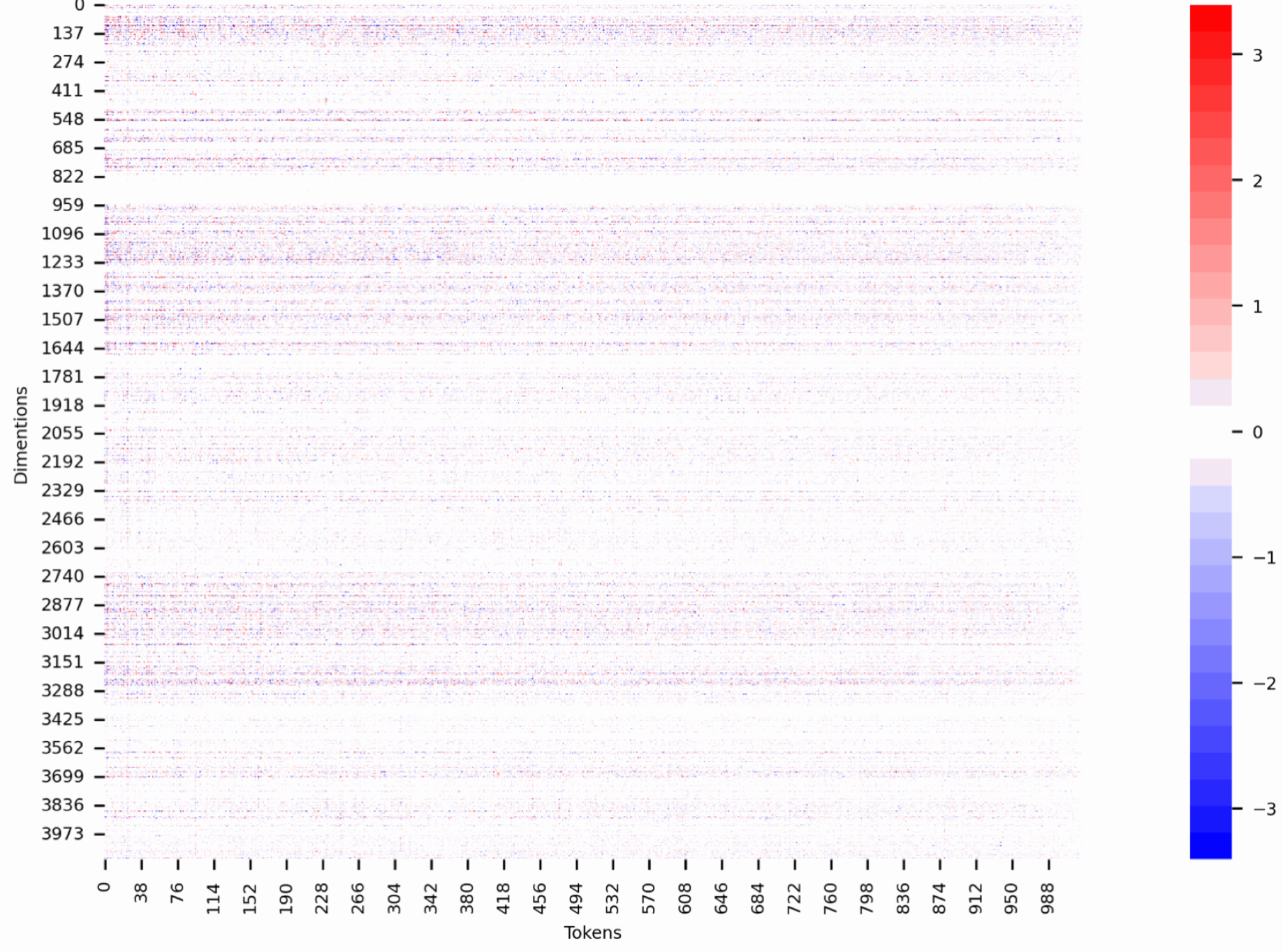}%
    \label{fig:channels_baseline_experts_backward}}
    \hfil
    \subfloat[PowLU, Backward]{\includegraphics[width=0.24\linewidth]{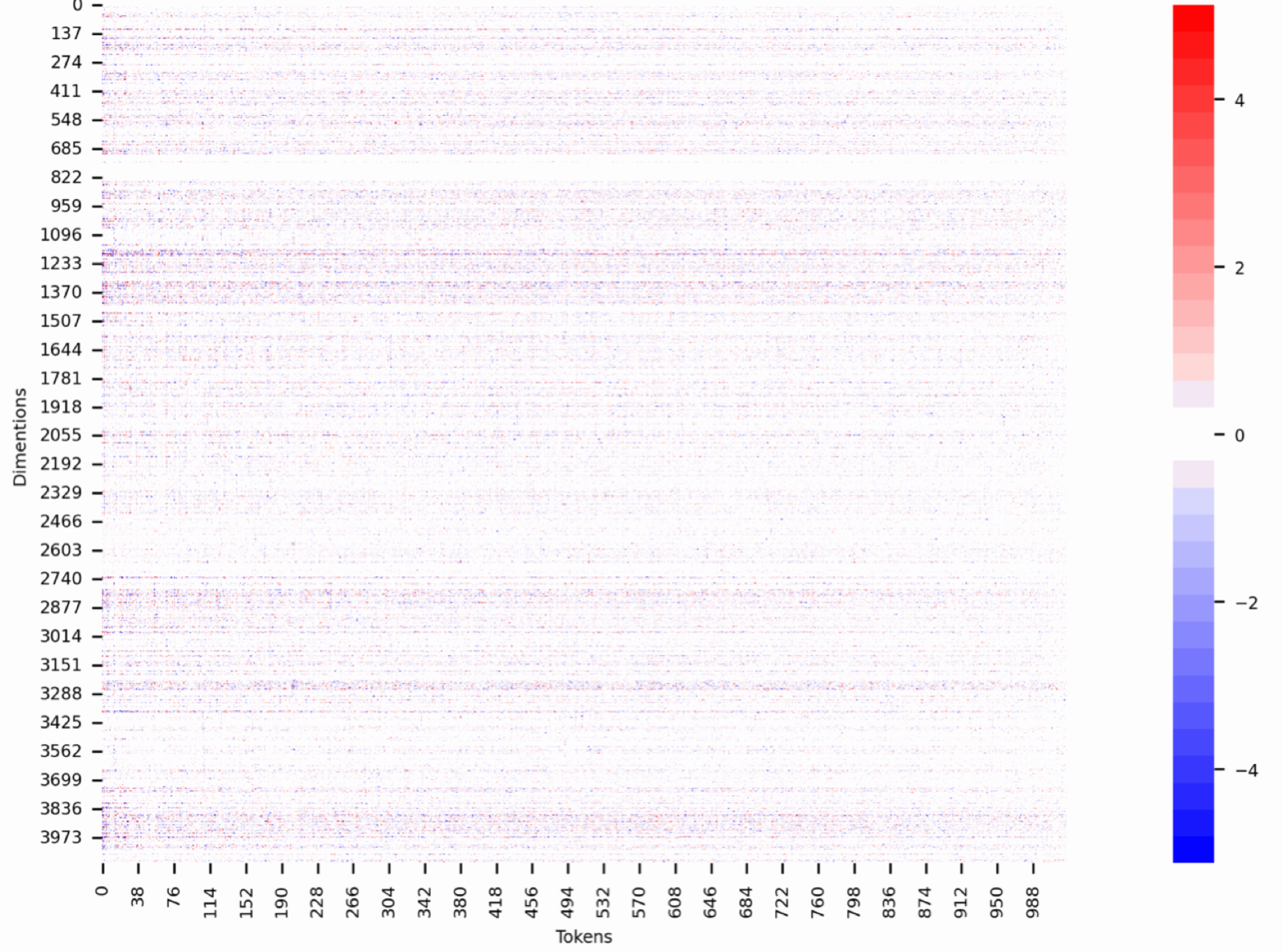}%
    \label{fig:channels_new_act_experts_backward}}
    \caption{Visualization of the outlier channels in the low-precision training of 7.9B MoE LLMs with SwiGLU activation function or PowLU activation function. The visualized layers are the linear layers in the experts of the LLM.}
    \label{fig:vis_channels_experts}
\end{figure*}

\begin{figure*}[ht]
    \centering
    \subfloat[SwiGLU, Forward]{\includegraphics[width=0.24\linewidth]{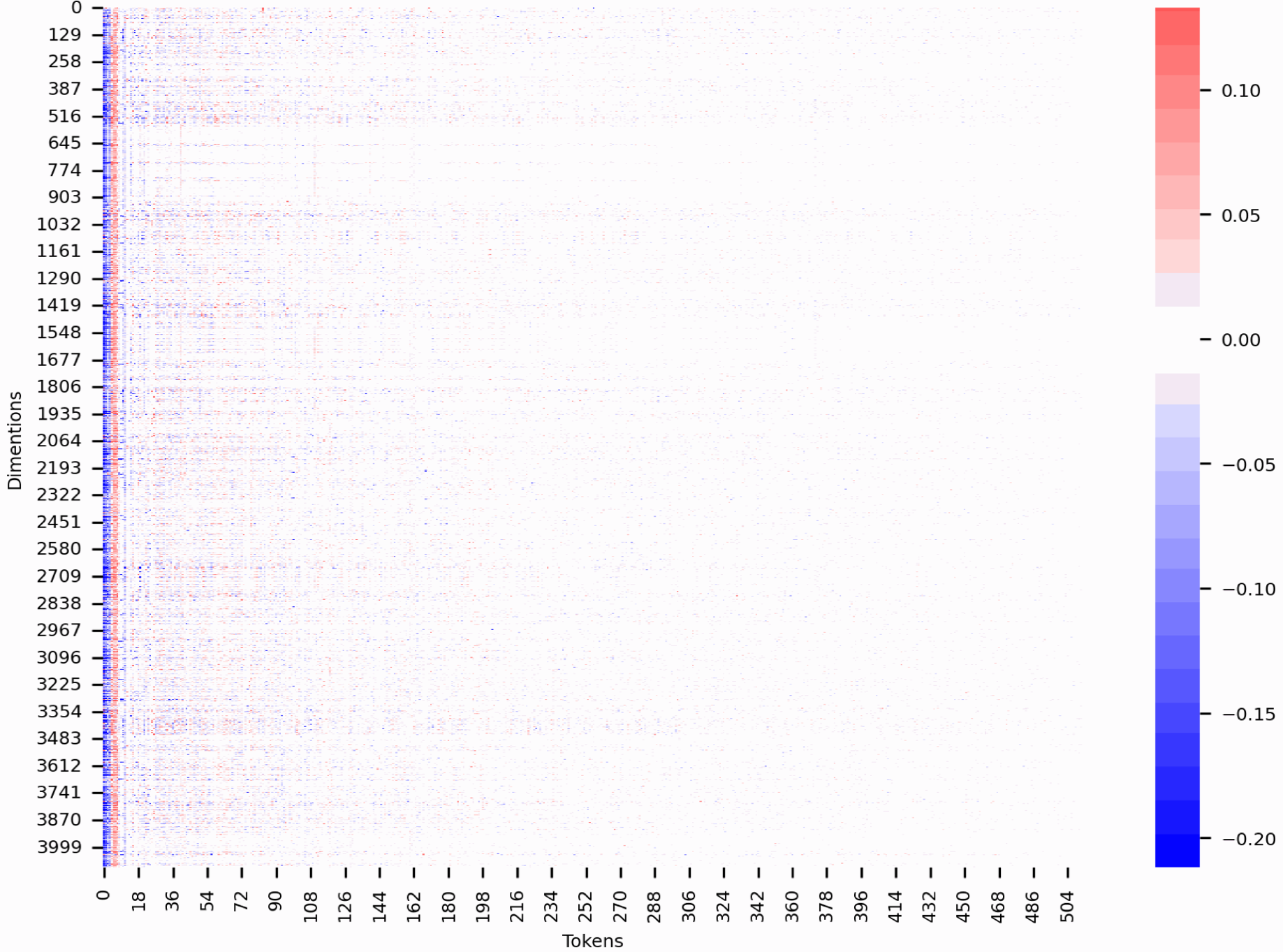}%
    \label{fig:channels_baseline_shared_experts_forward}}
    \hfil
    \subfloat[PowLU, Forward]{\includegraphics[width=0.24\linewidth]{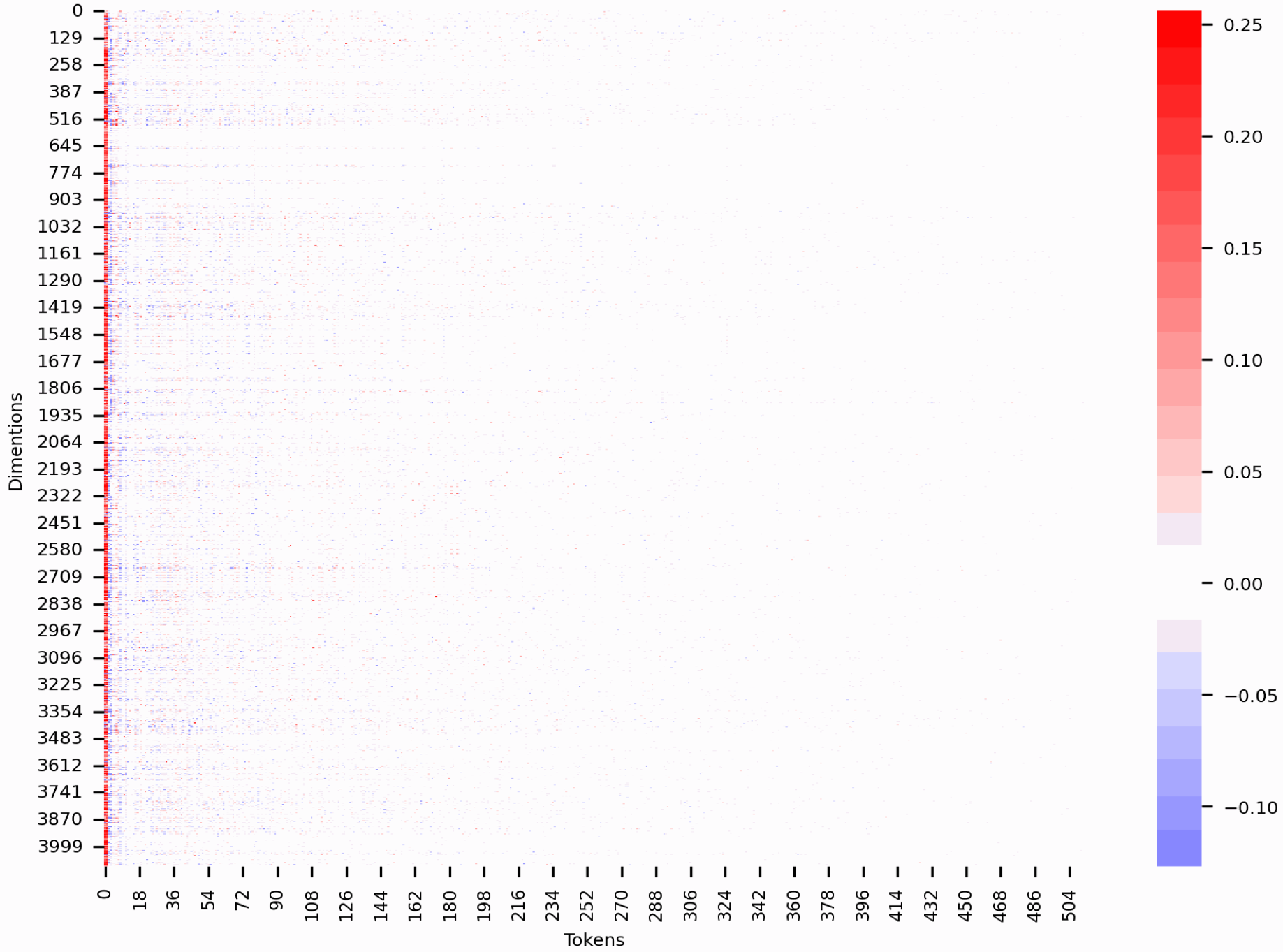}%
    \label{fig:channels_new_act_shared_experts_forward}}
    \hfil
    \subfloat[SwiGLU, Backward]{\includegraphics[width=0.24\linewidth]{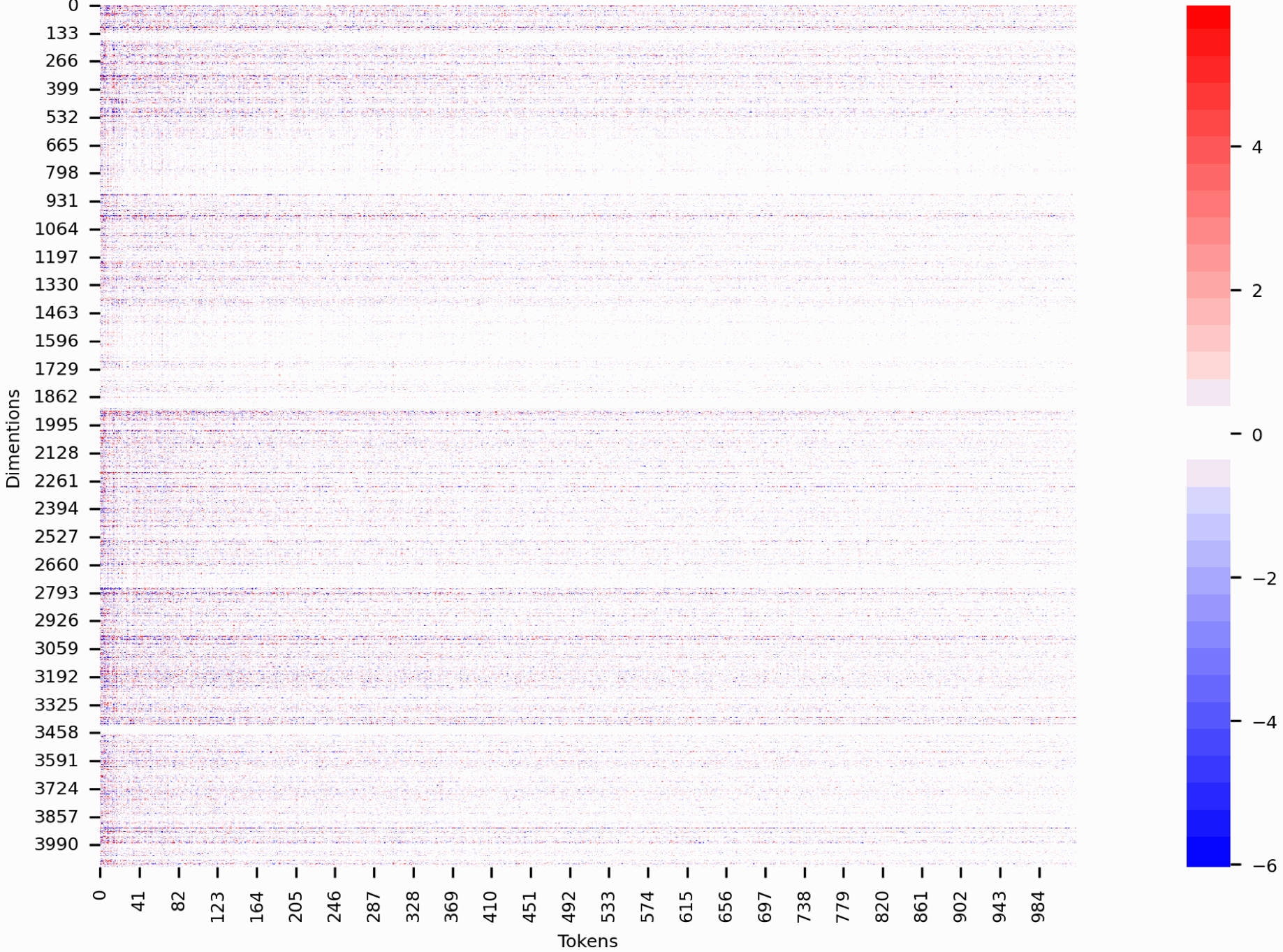}%
    \label{fig:channels_baseline_shared_experts_backward}}
    \hfil
    \subfloat[PowLU, Backward]{\includegraphics[width=0.24\linewidth]{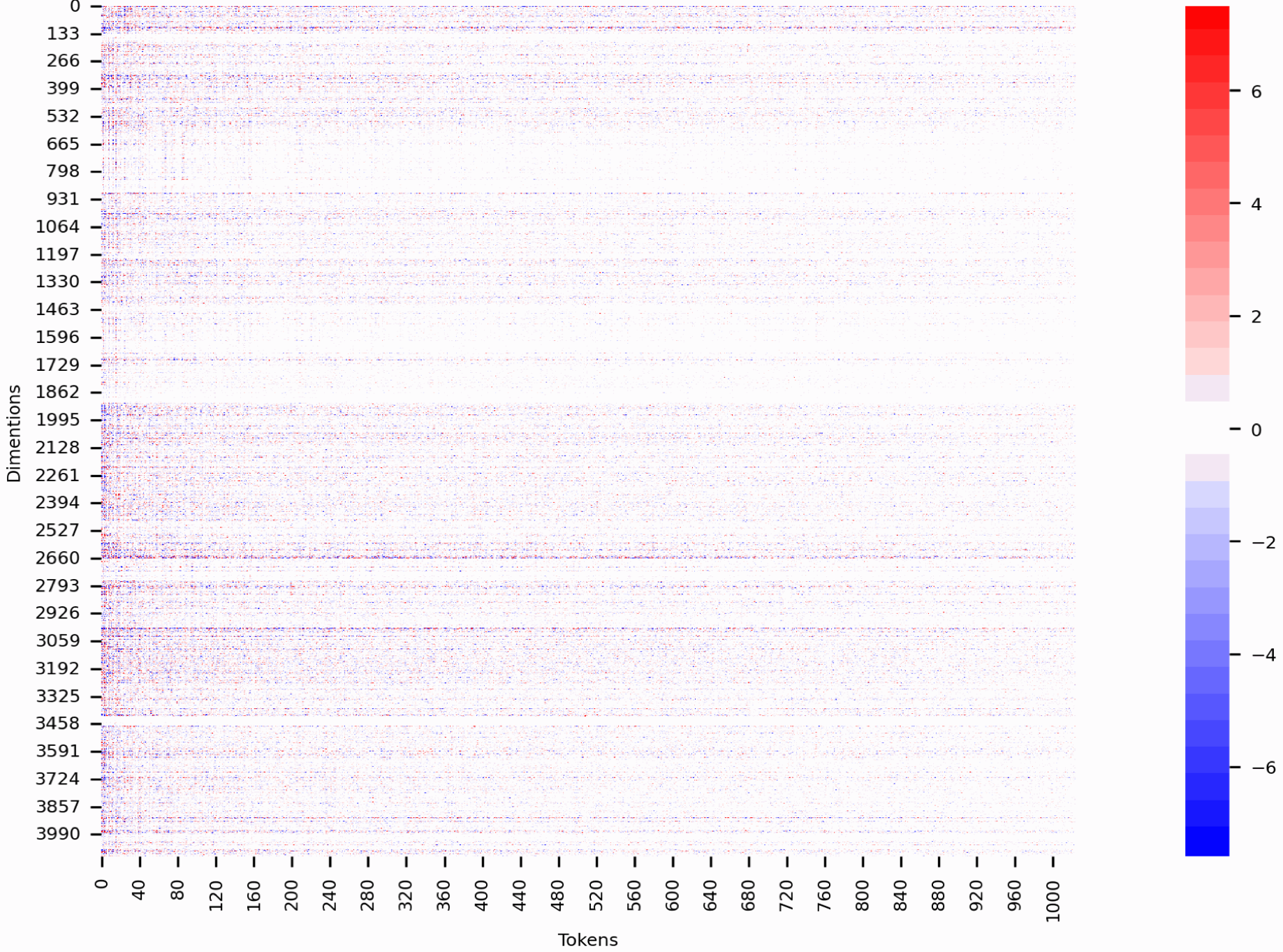}%
    \label{fig:channels_new_act_shared_experts_backward}}
    \caption{Visualization of the outlier channels in the low-precision training of 7.9B MoE LLMs with SwiGLU activation function or PowLU activation function. The visualized layers are the linear layers in the shared experts of the LLM.}
    \label{fig:vis_channels_shared_experts}
\end{figure*}

To further characterize outliers in tensors of experts and shared experts, we visualize the outlier channels (i.e., hidden dimension) of the same tensors as those in the numerical distribution analysis. For each tensor, we compute the $L_2$ norm along the channel axis and sort all channels in descending order by their norm magnitude to facilitate observation. The experimental results are presented in Figures~\ref{fig:vis_channels_experts} and \ref{fig:vis_channels_shared_experts}.

The visualizations in Figure~\ref{fig:vis_channels_experts} demonstrate the efficacy of the proposed PowLU activation function in suppressing outlier channels compared to the standard SwiGLU activation function. For the visualizations of SwiGLU in Figures~\ref{fig:channels_baseline_experts_forward} and \ref{fig:channels_baseline_experts_backward}, there is a prevalence of high-intensity red and blue regions, indicating that specific channels exhibit large magnitudes during both forward and backward propagation. In contrast, the corresponding visualizations for PowLU display a more uniform color distribution along with a compressed dynamic range. Moreover, we can make similar observations from Figure~\ref{fig:vis_channels_shared_experts}. The analysis above indicates that PowLU effectively mitigates the extreme outliers, resulting in smoother tensor distributions which are inherently more stable for the large-scale LLM training process.

\subsection{Ablation Studies}

We conduct ablation studies from two aspects, i.e., parameter studies for the hyperparameter $m$ and ablation studies for core components in the formulation of PowLU.

\begin{table*}[bp]
    \caption{Ablation studies for the hyperparameter $m$. The LLM is an MoE model with 47M activated parameters. The loss value is recorded after training with 29.8B tokens.}
    \label{tab:ablation_m}
    \centering
    \begin{tabular}{ccc}
    \toprule
    \textbf{Functions} & $m$ & \textbf{Loss} \\
    \midrule
    SwiGLU & N/A & 1.910 \\
    \midrule
    \multirow{3}{*}{PowLU} & 2 & 1.913 (+0.003) \\
    & 3 & 1.912 (+0.002) \\
    & 4 & 1.914 (+0.004) \\
    \bottomrule
    \end{tabular}
\end{table*}

\subsubsection{Parameter Studies} \label{subsubsec:parameter_study}

In the formulation of the PowLU activation function, there is a hyperparameter $m$ which can affect the degree of non-linearity when the input $x$ is small. In order to explore the influence of the hyperparameter $m$ on the performance of trained LLM, we conducted parameter studies for the hyperparameter $m$. Specifically, an MoE LLM with 47M activated parameters is used in the experiments, in order to ensure the training efficiency. Meanwhile, the hyperparameter $m$ is set to 2, 3, and 4 in the training process, and the training loss is compared with that of the SwiGLU baseline. Please note that the hyperparameter $m$ can be set continuously, we only choose 2, 3, and 4 in the experiments according to the distance between the curves of PowLU and SwiGLU activation functions. The experimental results are shown in Table~\ref{tab:ablation_m}.

According to the experimental results, the SwiGLU baseline achieves a loss of 1.910. In comparison, the PowLU function with $m = 3$ yielded the best performance among the tested configurations, achieving a loss of 1.912, which is only marginally higher than the baseline by +0.002. The other configurations, i.e., $m = 2$ and $m = 4$, resulted in slightly higher losses of 1.913 (+0.003) and 1.914 (+0.004), respectively. These results indicate that the effectiveness of PowLU activation function is not sensitive to the choice of the hyperparameter $m$, with $m = 3$ shown to be the optimal setting in this experimental setup.

\begin{figure*}[!tbp]
    \centering
    \subfloat[Activation Functions]{\includegraphics[width=0.49\linewidth]{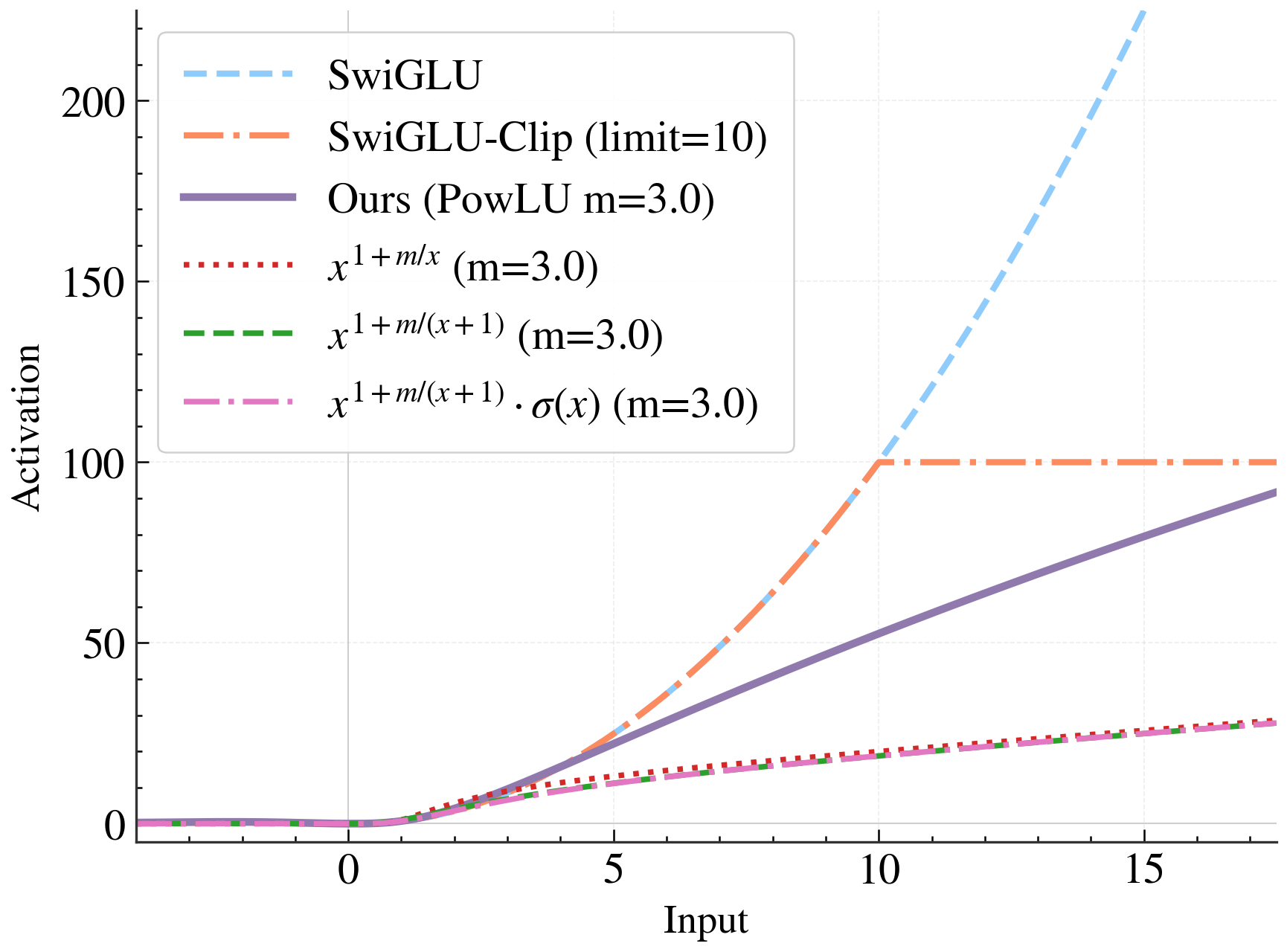}%
    \label{fig:ablation_functions}}
    \hfil
    \subfloat[First-Order Derivatives of Functions]{\includegraphics[width=0.49\linewidth]{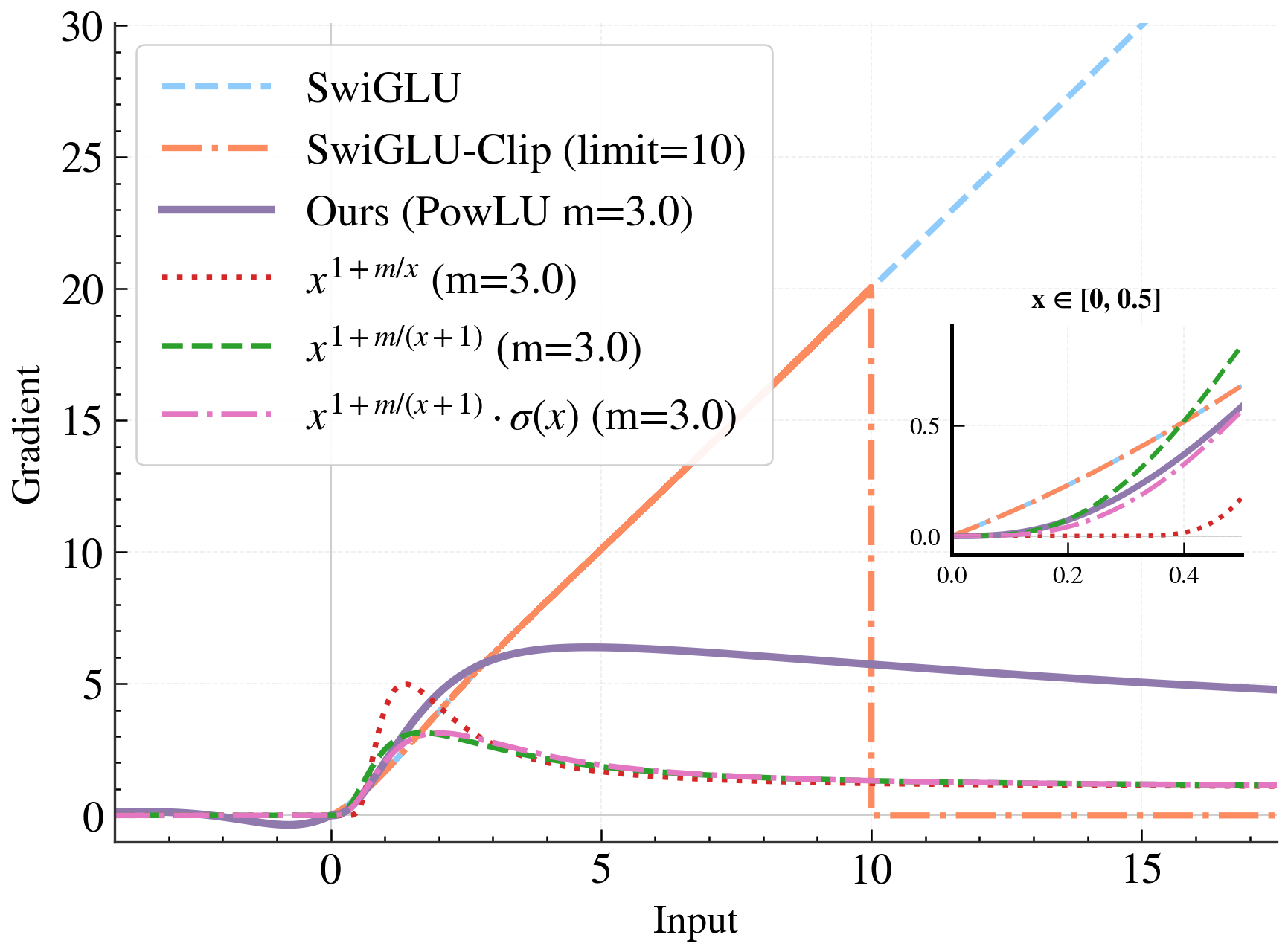}%
    \label{fig:ablation_derivations}}
    \caption{Visualizations of curves and first-order derivatives for different activation functions.}
    \label{fig:vis_ablation_functions}
\end{figure*}

\begin{figure*}[!tbp]
    \centering
    \includegraphics[width=\linewidth]{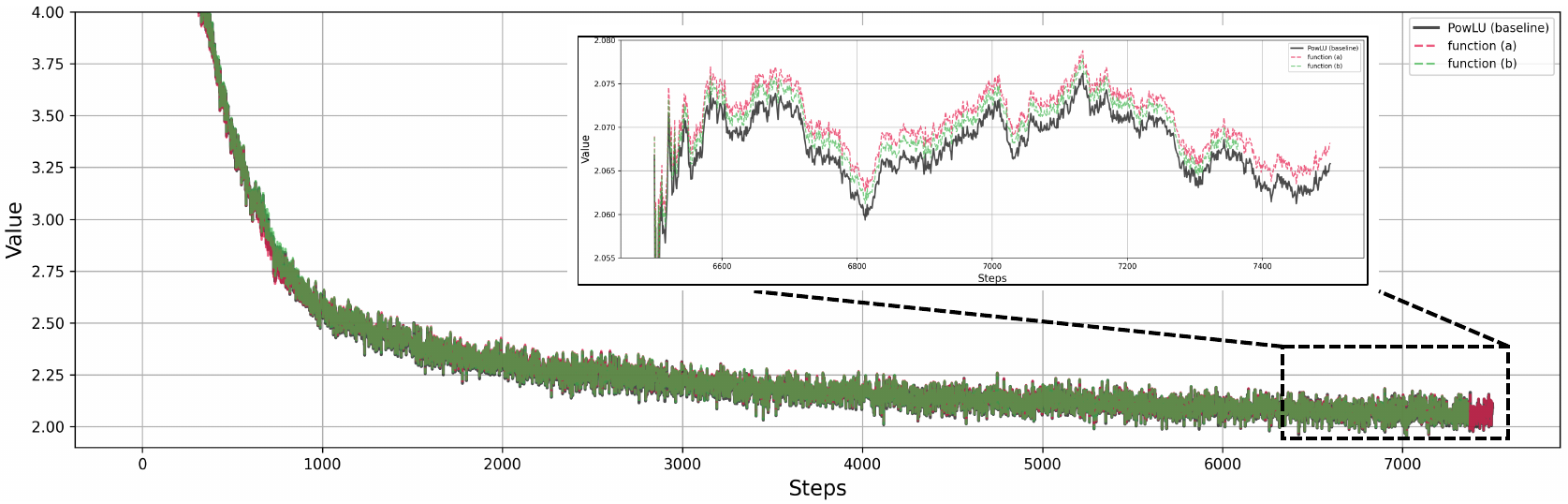}
    \caption{Visualizations for the loss difference between different activation functions and SwiGLU.}
    \label{fig:vis_loss_diff_functions}
\end{figure*}

\subsubsection{Ablation Studies for Core Components}

In order to validate the appropriateness of the activation function design, we conduct ablation studies in terms of the core components in PowLU. Specifically, the ablation study mainly focuses on the term $1$ of the denominator in the exponent, the sigmoid function, and the term $\sqrt{x}$ of the denominator in the exponent. To simulate different activation functions in the case of $x > 0$, we design three additional activation functions, i.e., $x^{1 + m / x}$, $x^{1 + m / (x + 1)}$, and $x^{1 + m / (x + 1)} \cdot \text{sigmoid}(x)$. In the $x \leq 0$ part, the formulation of these activation functions is the same as SwiGLU. The function curves and the corresponding first-order derivatives are presented in Figure~\ref{fig:vis_ablation_functions}.

As can be observed from Figure~\ref{fig:ablation_functions}, the growth rate of PowLU is between that of SwiGLU and the three functions designed when $x$ becomes larger. Moreover, the three functions designed have similar growth trends. As for the first-order derivatives in Figure~\ref{fig:ablation_derivations}, the derivatives of PowLU and three functions designed tend to plateau after rising. Meanwhile, the derivatives of PowLU are larger than that of the three designed functions. In addition, the subfigure on the right shows that the gradient of function $x^{1 + m / x}$ is almost invisible in interval $x \in (0, 0.5]$. This is mainly because the term $m / x$ approaches $+\infty$ as $x \rightarrow 0^{+}$. To this end, we introduce the term $1$ of the denominator in the exponent to avoid the above problem of numerical instability.

In addition, the training loss curves in terms of different activation functions are presented in Figure~\ref{fig:vis_loss_diff_functions}. Please note that the function (a) and function (b) in this figure refer to activation functions $x^{1 + m / (x + 1)}$ and $x^{1 + m / (\sqrt{x} + 1)} \cdot \text{sigmoid}(x)$, respectively. As can be seen from Figure~\ref{fig:vis_loss_diff_functions}, the proposed PowLU activation function achieves the lowest loss value compared to other two activation functions. These results support the validity of introducing terms $\sqrt{x}$ and $\text{sigmoid}(x)$ in the formulation of PowLU.

\section{Conclusion}

In this work, we identify the numerical instability caused by the commonly-used SwiGLU activation function in the LLM training. To tackle this issue, we propose the PowLU activation function as a solution. PowLU theoretically ensures non-linearity while constraining output ranges for large inputs via a rational power function, effectively mitigating outliers. Experimental results demonstrate that the PowLU-based LLM achieves competitive performance against the SwiGLU-based or SwiGLU-Clip-based LLMs across knowledge, reasoning, and code benchmarks. In addition, the analysis and ablation studies show that PowLU can effectively enhance the training stability of LLMs.

\bibliographystyle{assets/plainnat}
\bibliography{main}

\begin{thebibliography}{48}
\providecommand{\natexlab}[1]{#1}
\providecommand{\url}[1]{\texttt{#1}}
\expandafter\ifx\csname urlstyle\endcsname\relax
  \providecommand{\doi}[1]{doi: #1}\else
  \providecommand{\doi}{doi: \begingroup \urlstyle{rm}\Url}\fi

\bibitem[Agarwal et~al.(2025)Agarwal, Ahmad, Ai, Altman, Applebaum, Arbus, Arora, Bai, Baker, Bao, et~al.]{agarwal2025gpt}
Sandhini Agarwal, Lama Ahmad, Jason Ai, Sam Altman, Andy Applebaum, Edwin Arbus, Rahul~K Arora, Yu~Bai, Bowen Baker, Haiming Bao, et~al.
\newblock gpt-oss-120b \& gpt-oss-20b model card.
\newblock \emph{arXiv preprint arXiv:2508.10925}, 2025.

\bibitem[Bai(2022)]{bai2022relu}
Yuhan Bai.
\newblock {ReLU}-function and derived function review.
\newblock In \emph{SHS Web of Conferences}, volume 144, page 02006, 2022.

\bibitem[Chavan et~al.(2024)Chavan, Magazine, Kushwaha, Debbah, and Gupta]{chavan2024faster}
Arnav Chavan, Raghav Magazine, Shubham Kushwaha, M{\'e}rouane Debbah, and Deepak Gupta.
\newblock Faster and lighter {LLMs}: A survey on current challenges and way forward.
\newblock \emph{arXiv preprint arXiv:2402.01799}, 2024.

\bibitem[Chen et~al.(2021)Chen, Tworek, Jun, Yuan, Pinto, Kaplan, Edwards, Burda, Joseph, Brockman, et~al.]{chen2021evaluating}
Mark Chen, Jerry Tworek, Heewoo Jun, Qiming Yuan, Henrique Ponde De~Oliveira Pinto, Jared Kaplan, Harri Edwards, Yuri Burda, Nicholas Joseph, Greg Brockman, et~al.
\newblock Evaluating large language models trained on code.
\newblock \emph{arXiv preprint arXiv:2107.03374}, 2021.

\bibitem[Clark et~al.(2018)Clark, Cowhey, Etzioni, Khot, Sabharwal, Schoenick, and Tafjord]{clark2018think}
Peter Clark, Isaac Cowhey, Oren Etzioni, Tushar Khot, Ashish Sabharwal, Carissa Schoenick, and Oyvind Tafjord.
\newblock Think you have solved question answering? try {ARC}, the {AI2} reasoning challenge.
\newblock \emph{arXiv preprint arXiv:1803.05457}, 2018.

\bibitem[Cobbe et~al.(2021)Cobbe, Kosaraju, Bavarian, Chen, Jun, Kaiser, Plappert, Tworek, Hilton, Nakano, et~al.]{cobbe2021training}
Karl Cobbe, Vineet Kosaraju, Mohammad Bavarian, Mark Chen, Heewoo Jun, Lukasz Kaiser, Matthias Plappert, Jerry Tworek, Jacob Hilton, Reiichiro Nakano, et~al.
\newblock Training verifiers to solve math word problems.
\newblock \emph{arXiv preprint arXiv:2110.14168}, 2021.

\bibitem[Dauphin et~al.(2017)Dauphin, Fan, Auli, and Grangier]{dauphin2017language}
Yann~N Dauphin, Angela Fan, Michael Auli, and David Grangier.
\newblock Language modeling with gated convolutional networks.
\newblock In \emph{International Conference on Machine Learning}, pages 933--941. PMLR, 2017.

\bibitem[Devlin et~al.(2019)Devlin, Chang, Lee, and Toutanova]{devlin-etal-2019-bert}
Jacob Devlin, Ming-Wei Chang, Kenton Lee, and Kristina Toutanova.
\newblock {BERT}: Pre-training of deep bidirectional transformers for language understanding.
\newblock In \emph{Proceedings of the Conference of the North {A}merican Chapter of the Association for Computational Linguistics}, pages 4171--4186, 2019.

\bibitem[Du et~al.(2025)Du, Yao, Ma, Wang, Zheng, Zhu, Liu, Liang, Jin, Wei, Zheng, Deng, Guo, Jia, Jiang, Liao, Li, Li, Li, LI, Li, dehua ma, Ni, Que, Wang, Wen, Wu, Xing, Xu, Yang, Wang, Zhou, yuelin bai, Bu, chenglin cai, Chen, Chen, Chengtuo, Cheng, Ding, Huang, YUN, Li, Li, Li, Liang, Lin, Lin, Ma, Peng, Peng, Qi, Qiu, Qu, Quan, Tan, Wang, Wang, Wang, Wang, Wang, Xu, Yang, Yuan, Yue, Zhan, Zhang, Zhang, Zhang, Zhang, Zhang, Zhao, Zheng, ChenghuaZhong, Gao, Li, Liu, Liu, Liu, Ni, Peng, Qin, Su, Wang, Wang, Yang, Yang, Cao, Yue, Zhang, Zhou, Liu, Lin, Huang, and Zhang]{du2025supergpqa}
Xeron Du, Yifan Yao, Kaijing Ma, Bingli Wang, Tianyu Zheng, King Zhu, Minghao Liu, Yiming Liang, Xiaolong Jin, Zhenlin Wei, Chujie Zheng, Kaixin Deng, Shuyue Guo, Shian Jia, Sichao Jiang, Yiyan Liao, Rui Li, Qinrui Li, Sirun Li, Yizhi LI, Yunwen Li, dehua ma, Yuansheng Ni, Haoran Que, Qiyao Wang, Zhoufutu Wen, Siwei Wu, Tianshun Xing, Ming Xu, Zhenzhu Yang, Zekun~Moore Wang, Junting Zhou, yuelin bai, Xingyuan Bu, chenglin cai, Liang Chen, Yifan Chen, Cheng Chengtuo, Tianhao Cheng, Keyi Ding, Siming Huang, HUANG YUN, Yaoru Li, Yizhe Li, Zhaoqun Li, Tianhao Liang, Chengdong Lin, Hongquan Lin, Yinghao Ma, Z.Y. Peng, Zifan Peng, Qige Qi, Shi Qiu, Xingwei Qu, Shanghaoran Quan, Yizhou Tan, Zili Wang, Chenqing Wang, Hao Wang, Yiya Wang, Yubo Wang, Jiajun Xu, Kexin Yang, Ruibin Yuan, Yuanhao Yue, Tianyang Zhan, Chun Zhang, Jinyang Zhang, Xiyue Zhang, Owen~Xingjian Zhang, Yue Zhang, Yongchi Zhao, Xiangyu Zheng, ChenghuaZhong, Yang Gao, Zhoujun Li, Dayiheng Liu, Qian Liu, Tianyu Liu, Shiwen Ni, Junran Peng, Yujia Qin,
  Wenbo Su, Guoyin Wang, Shi Wang, Jian Yang, Min Yang, Meng Cao, Xiang Yue, Zhaoxiang Zhang, Wangchunshu Zhou, Jiaheng Liu, Qunshu Lin, Wenhao Huang, and Ge~Zhang.
\newblock Super{GPQA}: Scaling {LLM} evaluation across 285 graduate disciplines.
\newblock In \emph{The Thirty-ninth Annual Conference on Neural Information Processing Systems Datasets and Benchmarks Track}, 2025.
\newblock \url{https://openreview.net/forum?id=6WgflzYQpf}.

\bibitem[Grattafiori et~al.(2024)Grattafiori, Dubey, Jauhri, Pandey, Kadian, Al-Dahle, Letman, Mathur, Schelten, Vaughan, et~al.]{grattafiori2024llama}
Aaron Grattafiori, Abhimanyu Dubey, Abhinav Jauhri, Abhinav Pandey, Abhishek Kadian, Ahmad Al-Dahle, Aiesha Letman, Akhil Mathur, Alan Schelten, Alex Vaughan, et~al.
\newblock The {Llama} 3 herd of models.
\newblock \emph{arXiv preprint arXiv:2407.21783}, 2024.

\bibitem[Hao et~al.(2025)Hao, Guo, Shen, Luo, Hu, Wang, Yu, Wen, and Tao]{hao2025low}
Zhiwei Hao, Jianyuan Guo, Li~Shen, Yong Luo, Han Hu, Guoxia Wang, Dianhai Yu, Yonggang Wen, and Dacheng Tao.
\newblock Low-precision training of large language models: Methods, challenges, and opportunities.
\newblock \emph{arXiv preprint arXiv:2505.01043}, 2025.

\bibitem[Hendrycks and Gimpel(2016)]{hendrycks2016gaussian}
Dan Hendrycks and Kevin Gimpel.
\newblock Gaussian error linear units {(GELUs)}.
\newblock \emph{arXiv preprint arXiv:1606.08415}, 2016.

\bibitem[Hendrycks et~al.(2021{\natexlab{a}})Hendrycks, Burns, Basart, Zou, Mazeika, Song, and Steinhardt]{hendrycks2021measuring}
Dan Hendrycks, Collin Burns, Steven Basart, Andy Zou, Mantas Mazeika, Dawn Song, and Jacob Steinhardt.
\newblock Measuring massive multitask language understanding.
\newblock In \emph{International Conference on Learning Representations}, 2021{\natexlab{a}}.
\newblock \url{https://openreview.net/forum?id=d7KBjmI3GmQ}.

\bibitem[Hendrycks et~al.(2021{\natexlab{b}})Hendrycks, Burns, Kadavath, Arora, Basart, Tang, Song, and Steinhardt]{hendrycks2021math}
Dan Hendrycks, Collin Burns, Saurav Kadavath, Akul Arora, Steven Basart, Eric Tang, Dawn Song, and Jacob Steinhardt.
\newblock Measuring mathematical problem solving with the math dataset.
\newblock \emph{arXiv preprint arXiv:2103.03874}, 2021{\natexlab{b}}.

\bibitem[Huang et~al.(2025{\natexlab{a}})Huang, Hu, Zhang, Jin, Li, Shen, Chen, Liu, Wen, Wang, and Liu]{huang2025stablespam}
Tianjin Huang, Haotian Hu, Zhenyu Zhang, Gaojie Jin, Xiang Li, Li~Shen, Tianlong Chen, Lu~Liu, Qingsong Wen, Zhangyang Wang, and Shiwei Liu.
\newblock Stable-{SPAM}: How to train in 4-bit more stably than 16-bit {Adam}.
\newblock In \emph{First Workshop on Scalable Optimization for Efficient and Adaptive Foundation Models}, 2025{\natexlab{a}}.
\newblock \url{https://openreview.net/forum?id=Gk2pBIAMUl}.

\bibitem[Huang et~al.(2025{\natexlab{b}})Huang, Zhu, Jin, Liu, Wang, and Liu]{huang2025spam}
Tianjin Huang, Ziquan Zhu, Gaojie Jin, Lu~Liu, Zhangyang Wang, and Shiwei Liu.
\newblock {SPAM}: Spike-aware {Adam} with momentum reset for stable {LLM} training.
\newblock In \emph{International Conference on Learning Representations}, 2025{\natexlab{b}}.
\newblock \url{https://openreview.net/forum?id=L9eBxTCpQG}.

\bibitem[Huang et~al.(2023)Huang, Bai, Zhu, Zhang, Zhang, Su, Liu, Lv, Zhang, Fu, et~al.]{huang2023c}
Yuzhen Huang, Yuzhuo Bai, Zhihao Zhu, Junlei Zhang, Jinghan Zhang, Tangjun Su, Junteng Liu, Chuancheng Lv, Yikai Zhang, Yao Fu, et~al.
\newblock {C-Eval}: A multi-level multi-discipline chinese evaluation suite for foundation models.
\newblock \emph{Advances in Neural Information Processing Systems}, 36:\penalty0 62991--63010, 2023.

\bibitem[Joshi et~al.(2017)Joshi, Choi, Weld, and Zettlemoyer]{joshi2017triviaqa}
Mandar Joshi, Eunsol Choi, Daniel~S Weld, and Luke Zettlemoyer.
\newblock {TriviaQA}: A large scale distantly supervised challenge dataset for reading comprehension.
\newblock In \emph{Proceedings of the 55th Annual Meeting of the Association for Computational Linguistics (Volume 1: Long Papers)}, pages 1601--1611, 2017.

\bibitem[Kunc and Kl{\'e}ma(2024)]{kunc2024three}
Vladim{\'\i}r Kunc and Ji{\v{r}}{\'\i} Kl{\'e}ma.
\newblock Three decades of activations: A comprehensive survey of 400 activation functions for neural networks.
\newblock \emph{arXiv preprint arXiv:2402.09092}, 2024.

\bibitem[Lee et~al.(2024)Lee, Bae, Kim, Kwon, and Lee]{lee2024fp8}
Joonhyung Lee, Jeongin Bae, Byeongwook Kim, Se~Jung Kwon, and Dongsoo Lee.
\newblock To {FP8} and back again: Quantifying the effects of reducing precision on {LLM} training stability.
\newblock \emph{arXiv preprint arXiv:2405.18710}, 2024.

\bibitem[Li et~al.(2024)Li, Zhang, Koto, Yang, Zhao, Gong, Duan, and Baldwin]{li2024cmmlu}
Haonan Li, Yixuan Zhang, Fajri Koto, Yifei Yang, Hai Zhao, Yeyun Gong, Nan Duan, and Timothy Baldwin.
\newblock {CMMLU}: Measuring massive multitask language understanding in {Chinese}.
\newblock In \emph{Findings of the Association for Computational Linguistics: ACL 2024}, pages 11260--11285, 2024.

\bibitem[Minaee et~al.(2024)Minaee, Mikolov, Nikzad, Chenaghlu, Socher, Amatriain, and Gao]{minaee2024large}
Shervin Minaee, Tomas Mikolov, Narjes Nikzad, Meysam Chenaghlu, Richard Socher, Xavier Amatriain, and Jianfeng Gao.
\newblock Large language models: A survey.
\newblock \emph{arXiv preprint arXiv:2402.06196}, 2024.

\bibitem[Mirzadeh et~al.(2023)Mirzadeh, Alizadeh, Mehta, Del~Mundo, Tuzel, Samei, Rastegari, and Farajtabar]{mirzadeh2023relu}
Iman Mirzadeh, Keivan Alizadeh, Sachin Mehta, Carlo~C Del~Mundo, Oncel Tuzel, Golnoosh Samei, Mohammad Rastegari, and Mehrdad Farajtabar.
\newblock {ReLU} strikes back: Exploiting activation sparsity in large language models.
\newblock \emph{arXiv preprint arXiv:2310.04564}, 2023.

\bibitem[Molybog et~al.(2023)Molybog, Albert, Chen, DeVito, Esiobu, Goyal, Koura, Narang, Poulton, Silva, et~al.]{molybog2023theory}
Igor Molybog, Peter Albert, Moya Chen, Zachary DeVito, David Esiobu, Naman Goyal, Punit~Singh Koura, Sharan Narang, Andrew Poulton, Ruan Silva, et~al.
\newblock A theory on {Adam} instability in large-scale machine learning.
\newblock \emph{arXiv preprint arXiv:2304.09871}, 2023.

\bibitem[Nair and Hinton(2010)]{nair2010rectified}
Vinod Nair and Geoffrey~E Hinton.
\newblock Rectified linear units improve restricted boltzmann machines.
\newblock In \emph{International Conference on Machine Learning}, pages 807--814, 2010.

\bibitem[OpenAI(2024)]{wang2024mmmlu}
OpenAI.
\newblock Multilingual massive multitask language understanding {(MMMLU)}.
\newblock 2024.
\newblock \url{https://huggingface.co/datasets/openai/MMMLU}.

\bibitem[Pascanu et~al.(2013)Pascanu, Mikolov, and Bengio]{pascanu2013difficulty}
Razvan Pascanu, Tomas Mikolov, and Yoshua Bengio.
\newblock On the difficulty of training recurrent neural networks.
\newblock In \emph{International Conference on Machine Learning}, pages 1310--1318, 2013.

\bibitem[Radford et~al.(2018)Radford, Narasimhan, Salimans, Sutskever, et~al.]{radford2018improving}
Alec Radford, Karthik Narasimhan, Tim Salimans, Ilya Sutskever, et~al.
\newblock Improving language understanding by generative pre-training.
\newblock 2018.

\bibitem[Ramachandran et~al.(2017)Ramachandran, Zoph, and Le]{ramachandran2017searching}
Prajit Ramachandran, Barret Zoph, and Quoc~V Le.
\newblock Searching for activation functions.
\newblock \emph{arXiv preprint arXiv:1710.05941}, 2017.

\bibitem[Ramapuram et~al.(2025)Ramapuram, Danieli, Dhekane, Weers, Busbridge, Ablin, Likhomanenko, Digani, Gu, Shidani, and Webb]{ramapuram2025theory}
Jason Ramapuram, Federico Danieli, Eeshan~Gunesh Dhekane, Floris Weers, Dan Busbridge, Pierre Ablin, Tatiana Likhomanenko, Jagrit Digani, Zijin Gu, Amitis Shidani, and Russell Webb.
\newblock Theory, analysis, and best practices for sigmoid self-attention.
\newblock In \emph{International Conference on Learning Representations}, 2025.
\newblock \url{https://openreview.net/forum?id=Zhdhg6n2OG}.

\bibitem[Sakaguchi et~al.(2021)Sakaguchi, Bras, Bhagavatula, and Choi]{sakaguchi2021winogrande}
Keisuke Sakaguchi, Ronan~Le Bras, Chandra Bhagavatula, and Yejin Choi.
\newblock {WinoGrande}: An adversarial winograd schema challenge at scale.
\newblock \emph{Communications of the ACM}, 64\penalty0 (9):\penalty0 99--106, 2021.

\bibitem[Shazeer(2020)]{shazeer2020glu}
Noam Shazeer.
\newblock {GLU} variants improve transformer.
\newblock \emph{arXiv preprint arXiv:2002.05202}, 2020.

\bibitem[Shi et~al.(2023)Shi, Suzgun, Freitag, Wang, Srivats, Vosoughi, Chung, Tay, Ruder, Zhou, Das, and Wei]{shi2023language}
Freda Shi, Mirac Suzgun, Markus Freitag, Xuezhi Wang, Suraj Srivats, Soroush Vosoughi, Hyung~Won Chung, Yi~Tay, Sebastian Ruder, Denny Zhou, Dipanjan Das, and Jason Wei.
\newblock Language models are multilingual chain-of-thought reasoners.
\newblock In \emph{International Conference on Learning Representations}, 2023.
\newblock \url{https://openreview.net/forum?id=fR3wGCk-IXp}.

\bibitem[Singh et~al.(2025)Singh, Fry, Perelman, Tart, Ganesh, El-Kishky, McLaughlin, Low, Ostrow, Ananthram, et~al.]{singh2025openai}
Aaditya Singh, Adam Fry, Adam Perelman, Adam Tart, Adi Ganesh, Ahmed El-Kishky, Aidan McLaughlin, Aiden Low, AJ~Ostrow, Akhila Ananthram, et~al.
\newblock {OpenAI} {GPT-5} system card.
\newblock \emph{arXiv preprint arXiv:2601.03267}, 2025.

\bibitem[Suzgun et~al.(2023)Suzgun, Scales, Sch{\"a}rli, Gehrmann, Tay, Chung, Chowdhery, Le, Chi, Zhou, et~al.]{suzgun2023challenging}
Mirac Suzgun, Nathan Scales, Nathanael Sch{\"a}rli, Sebastian Gehrmann, Yi~Tay, Hyung~Won Chung, Aakanksha Chowdhery, Quoc Le, Ed~Chi, Denny Zhou, et~al.
\newblock Challenging big-bench tasks and whether chain-of-thought can solve them.
\newblock In \emph{Findings of the Association for Computational Linguistics: ACL 2023}, pages 13003--13051, 2023.

\bibitem[Takase et~al.(2025)Takase, Kiyono, Kobayashi, and Suzuki]{takase2025spike}
Sho Takase, Shun Kiyono, Sosuke Kobayashi, and Jun Suzuki.
\newblock Spike no more: Stabilizing the pre-training of large language models.
\newblock In \emph{Second Conference on Language Modeling}, 2025.
\newblock \url{https://openreview.net/forum?id=52YBEzcI0l}.

\bibitem[Team et~al.(2023)Team, Anil, Borgeaud, Alayrac, Yu, Soricut, Schalkwyk, Dai, Hauth, Millican, et~al.]{team2023gemini}
Gemini Team, Rohan Anil, Sebastian Borgeaud, Jean-Baptiste Alayrac, Jiahui Yu, Radu Soricut, Johan Schalkwyk, Andrew~M Dai, Anja Hauth, Katie Millican, et~al.
\newblock Gemini: a family of highly capable multimodal models.
\newblock \emph{arXiv preprint arXiv:2312.11805}, 2023.

\bibitem[Team et~al.(2026)Team, Chen, Zhang, Su, Xu, Pan, Wang, Wang, Chen, Yin, et~al.]{team2026attention}
Kimi Team, Guangyu Chen, Yu~Zhang, Jianlin Su, Weixin Xu, Siyuan Pan, Yaoyu Wang, Yucheng Wang, Guanduo Chen, Bohong Yin, et~al.
\newblock Attention residuals.
\newblock \emph{arXiv preprint arXiv:2603.15031}, 2026.

\bibitem[Team et~al.(2025)Team, Li, Liu, Hu, Li, Zeng, Ye, Tang, Tian, Huang, et~al.]{team2025every}
Ling Team, Ang Li, Ben Liu, Binbin Hu, Bing Li, Bingwei Zeng, Borui Ye, Caizhi Tang, Changxin Tian, Chao Huang, et~al.
\newblock Every activation boosted: Scaling general reasoner to 1 trillion open language foundation.
\newblock \emph{arXiv preprint arXiv:2510.22115}, 2025.

\bibitem[Vaswani et~al.(2017)Vaswani, Shazeer, Parmar, Uszkoreit, Jones, Gomez, Kaiser, and Polosukhin]{vaswani2017attention}
Ashish Vaswani, Noam Shazeer, Niki Parmar, Jakob Uszkoreit, Llion Jones, Aidan~N Gomez, {\L}ukasz Kaiser, and Illia Polosukhin.
\newblock Attention is all you need.
\newblock \emph{Advances in Neural Information Processing Systems}, 30, 2017.

\bibitem[Wang et~al.(2024)Wang, Ma, Zhang, Ni, Chandra, Guo, Ren, Arulraj, He, Jiang, et~al.]{wang2024mmlu}
Yubo Wang, Xueguang Ma, Ge~Zhang, Yuansheng Ni, Abhranil Chandra, Shiguang Guo, Weiming Ren, Aaran Arulraj, Xuan He, Ziyan Jiang, et~al.
\newblock {MMLU-Pro}: A more robust and challenging multi-task language understanding benchmark.
\newblock \emph{Advances in Neural Information Processing Systems}, 37:\penalty0 95266--95290, 2024.

\bibitem[Wei et~al.(2023)Wei, Luan, Liu, Dong, and Wang]{wei2023cmath}
Tianwen Wei, Jian Luan, Wei Liu, Shuang Dong, and Bin Wang.
\newblock {CMATH}: Can your language model pass chinese elementary school math test?
\newblock \emph{arXiv preprint arXiv:2306.16636}, 2023.

\bibitem[Wei et~al.(2024)Wei, Moalla, Pascanu, and Gulcehre]{wei2024building}
Xiuying Wei, Skander Moalla, Razvan Pascanu, and Caglar Gulcehre.
\newblock Building on efficient foundations: Effective training of {LLMs} with structured feedforward layers.
\newblock \emph{Advances in Neural Information Processing Systems}, 37:\penalty0 4689--4717, 2024.

\bibitem[Wortsman et~al.(2024)Wortsman, Liu, Xiao, Everett, Alemi, Adlam, Co-Reyes, Gur, Kumar, Novak, Pennington, Sohl-Dickstein, Xu, Lee, Gilmer, and Kornblith]{wortsman2024smallscale}
Mitchell Wortsman, Peter~J Liu, Lechao Xiao, Katie~E Everett, Alexander~A Alemi, Ben Adlam, John~D Co-Reyes, Izzeddin Gur, Abhishek Kumar, Roman Novak, Jeffrey Pennington, Jascha Sohl-Dickstein, Kelvin Xu, Jaehoon Lee, Justin Gilmer, and Simon Kornblith.
\newblock Small-scale proxies for large-scale transformer training instabilities.
\newblock In \emph{International Conference on Learning Representations}, 2024.
\newblock \url{https://openreview.net/forum?id=d8w0pmvXbZ}.

\bibitem[Yang et~al.(2025)Yang, Li, Yang, Zhang, Hui, Zheng, Yu, Gao, Huang, Lv, et~al.]{yang2025qwen3}
An~Yang, Anfeng Li, Baosong Yang, Beichen Zhang, Binyuan Hui, Bo~Zheng, Bowen Yu, Chang Gao, Chengen Huang, Chenxu Lv, et~al.
\newblock Qwen3 technical report.
\newblock \emph{arXiv preprint arXiv:2505.09388}, 2025.

\bibitem[Zellers et~al.(2019)Zellers, Holtzman, Bisk, Farhadi, and Choi]{zellers2019hellaswag}
Rowan Zellers, Ari Holtzman, Yonatan Bisk, Ali Farhadi, and Yejin Choi.
\newblock {HellaSwag}: Can a machine really finish your sentence?
\newblock In \emph{Proceedings of the 57th Annual Meeting of the Association for Computational Linguistics}, pages 4791--4800, 2019.

\bibitem[Zhang et~al.(2024)Zhang, Liu, Cherry, and Firat]{zhang2024when}
Biao Zhang, Zhongtao Liu, Colin Cherry, and Orhan Firat.
\newblock When scaling meets {LLM} finetuning: The effect of data, model and finetuning method.
\newblock In \emph{International Conference on Learning Representations}, 2024.
\newblock \url{https://openreview.net/forum?id=5HCnKDeTws}.

\bibitem[Zhong et~al.(2024)Zhong, Cui, Guo, Liang, Lu, Wang, Saied, Chen, and Duan]{zhong2024agieval}
Wanjun Zhong, Ruixiang Cui, Yiduo Guo, Yaobo Liang, Shuai Lu, Yanlin Wang, Amin Saied, Weizhu Chen, and Nan Duan.
\newblock {AGIEval}: A human-centric benchmark for evaluating foundation models.
\newblock In \emph{Findings of the Association for Computational Linguistics: NAACL 2024}, pages 2299--2314, 2024.

\end{thebibliography}

\appendix

\section*{Appendix}

\section{Theoretical Analysis}

\subsection{Monotonicity}\label{subsec:monotonicity}

In this subsection, we analyze the monotonicity of PowLU when $x > 0$. For ease of representation, we denote the PowLU activation function as $f(x) = x^{1 + m / (\sqrt{x} + 1)} \sigma(x)$ in the case of $x > 0$. In this expression, $\sigma(x) = 1 / (1 + e^{-x})$ denotes the sigmoid function. Because the function $f(x)$ satisfies $f(x) > 0$, analyzing the monotonicity of $f(x)$ is equivalent to analyzing the monotonicity of $g(x) = \ln f(x)$. The specific form of $g(x)$ is shown in Eq.~(\ref{eq:gx}).
\begin{equation}\label{eq:gx}
    g(x) = \left( 1 + \frac{m}{\sqrt{x} + 1} \right) \ln x + \ln \sigma(x)
\end{equation}
Let $t = \sqrt{x} > 0$, the first-order derivative of $g(x)$ can be calculated as Eq.~(\ref{eq:dgx}):
\begin{equation}\label{eq:dgx}
    g'(x) = \frac{(t + 1)^2 + m \cdot \phi(t)}{t^2(t + 1)^2} + \frac{1}{1 + e^{t^2}}
\end{equation}
where $\phi(t) = t + 1 - t \ln t$.

Based on the above transformation process, we further analyze the properties of the auxiliary function $\phi(t)$. The first-order derivative of $\phi(t)$ can be written as $\phi'(t) = -\ln t$. According to the expression of $\phi'(t)$, we can know that $\phi(t)$ is monotonically increasing when $0 < t < 1$ and monotonically decreasing when $t > 1$. When $t = 1$, $\phi(t)$ achieves the global maximum value $\phi(1) = 2$. Furthermore, because $\lim_{t \rightarrow 0^{+}} \phi(t) = 1$, there exists a unique zero point $t_0 \approx 3.59$ of $\phi(t)$. Obviously, the function $\phi(t)$ satisfies $\phi(t) < 0$ when $t > t_0$ and $\phi(t) > 0$ when $0 < t < t_0$.

According to the above analysis, we analyze the monotonicity of $g(x)$ in the case of $m > 0$. When $t$ satisfies $0 < t \leq t_0$, function $\phi(t)$ satisfies $\phi(t) \geq 0$, thus we have Eq.~(\ref{eq:gx_derivation}) based on the scaling method.
\begin{equation}\label{eq:gx_derivation}
g'(x) \geq \frac{(t + 1)^2}{t^2 (t + 1)^2} + \frac{1}{1 + e^{t^2}} = \frac{1}{t^2} + \frac{1}{1 + e^{t^2}} > 0
\end{equation}
Furthermore, when $t > t_0$, the function $\phi(t)$ satisfies $\phi(t) < 0$. Because $1 / (1 + e^{t^2}) > 0$, a sufficient condition for $g'(x) > 0$ is $(t + 1)^2 + m \phi(t) \geq 0$. This condition can be further transformed as shown in Eq.~(\ref{eq:trans_scale}).
\begin{equation}\label{eq:trans_scale}
    (t + 1)^2 + m \phi(t) \geq 0 \Leftrightarrow m \leq \frac{(t + 1)^2}{t \ln t - t - 1} \equiv M(t)
\end{equation}
Then, we find the minimum value by differentiating $M(t)$, and we can know that the stationary point of $M(t)$ satisfies $\ln t = 2 + 4 / (t - 1)$. The solution $t^{*}$ to the transcendental equation is $t^{*} \approx 11.02$, thus the minimum value of $M(t)$ is approximately equal to 10.02. Therefore, the upper bound of $m$ is approximately equal to 10.02. In summary, PowLU is monotonically increasing when the hyperparameter $m$ satisfies $0 < m < 10$.

\subsection{Non-Linearity}

We analyze the non-linearity from two cases, i.e., $x > 0$ and $x \leq 0$. In terms of the $x > 0$ part, the PowLU activation function is a product of $x^{1 + m / (\sqrt{x} + 1)}$ and the sigmoid function. The sigmoid function is non-linear~\citep{ramapuram2025theory}. Furthermore, the exponent $1 + m / (\sqrt{x} + 1)$ is a function of $x$, making the function $x^{1 + m / (\sqrt{x} + 1)}$ non-linear as well. As a result, the product of two non-linear functions remains non-linear. In terms of the $x \leq 0$ part, the PowLU function is a product of a quadratic function $x^2$ and the sigmoid function. Similarly, this is a product of two non-linear functions and is also non-linear. According to the above analysis, the PowLU activation function is a non-linear function.

\end{document}